\documentclass[nohyperref]{article}

\usepackage{microtype}
\usepackage{graphicx}
\usepackage{subfigure}
\usepackage{booktabs} 

\usepackage{hyperref}

\usepackage[accepted]{icml2022_env/icml2022}

\usepackage{amsmath}
\usepackage{amssymb}
\usepackage{mathtools}
\usepackage{amsthm}

\usepackage[capitalize,noabbrev]{cleveref}

\theoremstyle{plain}

\theoremstyle{definition}

\theoremstyle{remark}

\usepackage[textsize=tiny]{todonotes}

\icmltitlerunning{Mitigating Gender Bias in Face Recognition}

\usepackage{bm}
\usepackage{amsfonts}
\newcommand{\vect}[1]{{\bm{#1}}}

\usepackage{multirow}
\usepackage{caption}

\begin{document}

\twocolumn[
\icmltitle{ Mitigating Gender Bias in Face Recognition \\
Using the von Mises-Fisher Mixture Model
}



\icmlsetsymbol{equal}{*}

\begin{icmlauthorlist}
\icmlauthor{Jean-Rémy Conti}{equal,telecom,idemia}
\icmlauthor{Nathan Noiry}{equal,telecom}
\icmlauthor{Vincent Despiegel}{idemia}
\icmlauthor{Stéphane Gentric}{idemia}
\icmlauthor{Stéphan Clémençon}{telecom}
\end{icmlauthorlist}

\icmlaffiliation{telecom}{LTCI, Télécom Paris, Institut Polytechnique de Paris}
\icmlaffiliation{idemia}{Idemia}

\icmlcorrespondingauthor{Jean-Rémy Conti}{jean-remy.conti@telecom-paris.fr}
\icmlcorrespondingauthor{Nathan Noiry}{nathan.noiry@gmail.com}

\icmlkeywords{Machine Learning, ICML}

\vskip 0.3in
]



\printAffiliationsAndNotice{\icmlEqualContribution} 

\begin{abstract}
In spite of the high performance and reliability of deep learning algorithms in a wide range of everyday applications, many investigations tend to show that a lot of models exhibit biases, discriminating against specific subgroups of the population (\textit{e.g.} gender, ethnicity). This urges the practitioner to develop fair systems with a uniform/comparable performance across sensitive groups. In this work, we investigate the gender bias of deep Face Recognition networks. In order to measure this bias, we introduce two new metrics, $\mathrm{BFAR}$ and $\mathrm{BFRR}$, that better reflect the inherent deployment needs of Face Recognition systems. Motivated by geometric considerations, we mitigate gender bias through a new post-processing methodology which transforms the deep embeddings of a pre-trained model to give more representation power to discriminated subgroups. It consists in training a shallow neural network by minimizing a {\it Fair von Mises-Fisher loss} whose hyperparameters account for the intra-class variance of each gender. Interestingly, we empirically observe that these hyperparameters are correlated with our fairness metrics. In fact, extensive numerical experiments on a variety of datasets show that a careful selection significantly reduces gender bias. The code used for the experiments can be found at \url{https://github.com/JRConti/EthicalModule_vMF}.


\end{abstract}

\section{Introduction}







In the past few years, Face Recognition (FR) systems have reached extremely high levels of performance, paving the way to a broader range of applications, where the reliability levels were previously prohibitive to consider automation. This is mainly due to the adoption of deep learning techniques in computer vision since the famous breakthrough of \cite{krizhevsky2012imagenet}. The increasing use of deep FR systems has however raised concerns as any technological flaw could have a strong societal impact. Besides recent punctual events\footnote{See for instance the \href{https://www.aclu.org/blog/privacy-technology/surveillance-technologies/amazons-face-recognition-falsely-matched-28}{study} conducted by the American Civil Liberties Union.} that received significant media coverage, the academic community has studied the bias of FR systems for many years (dating back at least to \cite{nist_report_2002} who investigated the racial bias of non-deep FR algorithms). In \cite{bias_old_effects} three sources of biases are identified: race (understood as biological attributes such as skin color), age and gender (available gender labels from FR datasets are males and females). The National Institute of Standards and Technology \cite{nist_report_2019} conducted a thorough analysis of the performance of several FR algorithms depending on these attributes and revealed high disparities. For instance, some of the top state-of-the-art algorithms in absolute performance have more than seven times more false acceptances for females than for males. In this paper, we introduce a novel methodology to mitigate gender bias for FR. Though focusing on a single source of bias has obvious limitations regarding intersectional effects \cite{gender_shades}, it is a first step to gain insights into the mechanisms at work, before turning to more complex situations. Actually, the method promoted in this paper, much more general than the application considered here, could possibly alleviate many other types of bias. This will be the subject of a future work.

The topic corresponding to the study of different types of bias and to the elaboration of methods to alleviate them is referred to as {\it fairness} in machine learning, which has received increasing attention in recent years, see \textit{e.g.} \cite{survey_fairness_ML}, \cite{caton2020fairness}, \cite{fairness_DL}. Roughly speaking, achieving fairness means learning a decision rule that does not mistreat some predefined subgroups, while still exhibiting a good predictive performance on the overall population: in general, a trade-off has to be found between fair treatment and pure accuracy\footnote{This dichotomy somewhat simplifies the problem since an increase in accuracy could also lead to a better treatment of each subgroup of the population.}. In this regard, one needs to carefully define what will be the relevant {\it fairness metric}. From a theoretical viewpoint, several ones have been introduced, see \textit{e.g.} \cite{garg2020fairness} or \cite{castelnovo2021zoo} among others, depending on how the concept of equity of treatment is understood. In practice, these very refined notions can be inadequate, as they ignore specific use case issues, and one thus needs to adapt them carefully. This is particularly the case in FR, where high security standards cannot be negotiated. The goal of this article is twofold: novel fairness metrics, relevant in FR applications in particular, are introduced at length and empirically shown to have room for improvement by means of appropriate/flexible representation models.

{\bf Contribution 1.} We propose two new metrics, $\mathrm{BFAR}$ and $\mathrm{BFRR}$, that incorporate the needs for both security and fairness (see section \ref{subsec:fairness}). More precisely, the $\mathrm{BFAR}$ (resp. $\mathrm{BFRR})$ metric accounts for the disparity between false acceptance (resp. rejection) rates between subgroups of interest, computed at an operating point such that each subgroup has a false acceptance rate lower than a false acceptance level of reference.

It turns out that state-of-the-art FR networks ({\it e.g.} ArcFace \cite{arcface}) exhibit poor fairness performance w.r.t. gender, both in terms of $\mathrm{BFAR}$ and $\mathrm{BFRR}$. Different strategies could be considered to alleviate this gender bias:  pre-, in- and post-processsing methods \cite{caton2020fairness}, depending on whether the practitioner ``fairness" intervention occurs before, during or after the training phase. The first one, pre-processing, is not well suited for FR purposes as shown in \cite{gender_balanced_data}, while the second one, in-processing, has the major drawback to require a full retraining of a deep neural network. This encouraged us to design a post-processing method so as to mitigate gender bias of pre-trained FR models.

In order to improve $\mathrm{BFAR}$ and $\mathrm{BFRR}$ disparities, we crucially rely on the geometric structure of the last layer of state-of-the-art FR neural networks. The latter is a set of embeddings lying on a \textit{hypersphere}. Those embeddings are obtained through two concurrent mechanisms at work during the learning process: {\it (i)} repel images of different identities and {\it (ii)} bring together images of a same identity. 

{\bf Contribution 2.} We set a von Mises-Fisher statistical mixture model on the last layer representation, which corresponds to a mixture of gaussian random variables conditioned to live on the hypersphere. Based on the maximum likelihood of this model, we introduce a new loss we call {\it Fair von Mises-Fisher}, that we use to supervise the training of a shallow neural network we call {\it Ethical Module}. Taking the variance parameters as hyperparameters that depend on the gender, this flexible model is able to capture the two previously mentioned mechanisms of repulsion / attraction, which we show are at the origin of the biases in FR. Indeed, our experiments remarkably exhibit a substantial correlation between these hyperparameters and our fairness metrics $\mathrm{BFAR}$ and $\mathrm{BFRR}$, suggesting a hidden regularity captured by the model proposed. More precisely, we identify some regions of hyperparameters' values that {\it (i)} significantly improve $\mathrm{BFAR}$ while keeping a reasonable performance but degrading $\mathrm{BFRR}$, {\it (ii)} significantly improve $\mathrm{BFRR}$ while keeping a reasonable performance but degrading $\mathrm{BFAR}$ and {\it (iii)} improve both $\mathrm{BFAR}$ and $\mathrm{BFRR}$ at the cost of little performance degradation. This third case actually achieves state-of-the-art results in terms of post-processing methods for gender bias mitigation in FR.

\begin{figure}[h]
    \centering
    \includegraphics[scale=0.45]{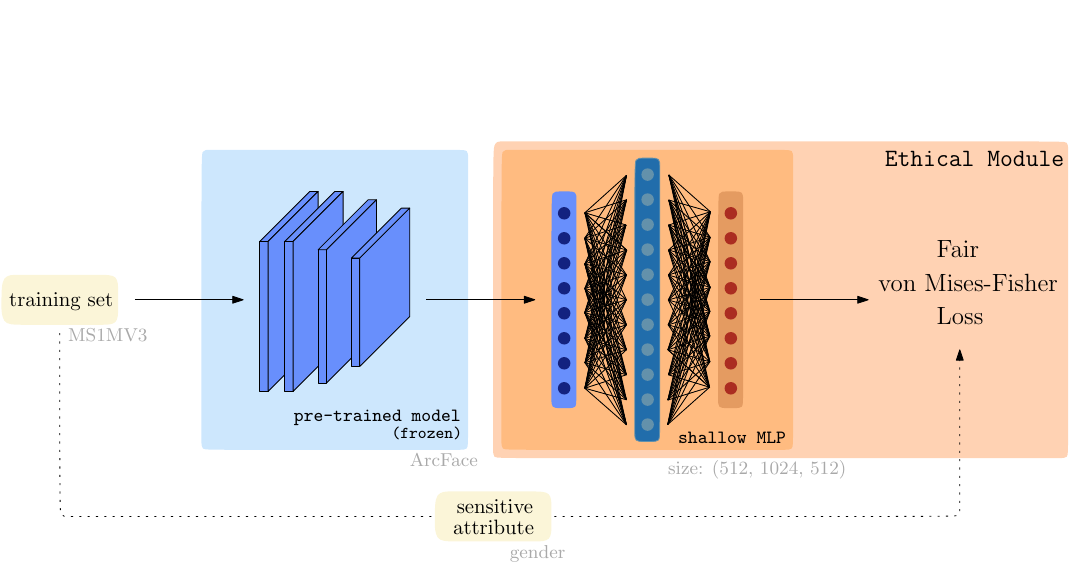}
    \caption{Illustration of the Ethical Module methodology. In gray: our experiment choices.
    }
    \label{fig:ethical_module}
\end{figure}

Besides a simple architecture and a fast training (few hours), the {\it Ethical Module} enjoys several benefits we would like to highlight. 

{\bf Taking advantage of foundation models.} In the recent survey \cite{bommasani2021opportunities}, the authors judiciously point out a change of paradigm in deep learning: very efficient pre-trained models with billions of parameters they call {\it foundation models} are at our disposal such as BERT \cite{devlin2018bert} in NLP or ArcFace \cite{arcface} in FR. Many works rely on these powerful models and fine tune them, inheriting from both their strengths and weaknesses such as their biases. Hence the need to focus on methods to improve the fairness of foundation models: our method is in line with this approach.

    

{\bf No sensitive attribute used during deployment.} Though the Ethical Module requires access to the sensitive label during its training phase, this label ({\it e.g.} gender) is not needed anymore, once the training is completed. This is compliant with the EU jurisdiction that forbids the use of protected attributes for prediction rules.

{\bf Organization of the paper.} Section \ref{subsec:FR_survey} presents the widely spread usage of FR and its main challenges. It is followed by section \ref{subsec:fairness} where we discuss different fairness metrics that arise in FR and introduce two new ones we think are more relevant with regards to operational use cases. In section \ref{sec:vMF for eth mod}, we present the von Mises-Fisher loss that is used for the training of the Ethical Module and discuss its benefits. Finally, in section \ref{sec:expe}, we present at length our numerical experiments, which partly consist in learning an Ethical Module on the ArcFace model, pre-trained on the MS1MV3 dataset \cite{ms1m_retinaface}. Our results show that, remarkably, some specific choices of hyperparameters provide high performance and low fairness metrics both at the same time.

{\bf Related works.} The correction of bias in FR has been the subject of several recent papers. \cite{fairloss_RL} and \cite{mitigating_bias_RL} use reinforcement learning to learn fair decision rules but despite their mathematical relevance, such methods are computationally prohibitive. Another line of research followed by \cite{transfer_learning_FR}, \cite{RFW_2019} and \cite{huang2019deep} assumes that bias comes from the unbalanced nature of FR datasets and builds on imbalanced and transfer learning methods. Unfortunately, these methods do dot completely remove bias and it has been recently pointed out that balanced dataset are actually {\it not enough} to mitigate bias, as illustrated by \cite{gender_balanced_data} for gender bias, \cite{race_balanced_data} for racial bias and \cite{balanced_datasets_are_not_enough} for gender bias in face detection. \cite{debface}, \cite{gan_fairness_gender} and \cite{dhar2021pass} rely on adversarial methods that can reduce bias but are also known to be unstable and computationally expensive. All of the previously mentioned methods try to learn fair representations. In contrast, some other works do not affect the latent space but modify the decision rule instead: \cite{terhorst2020post} act on the score function whereas \cite{salvador2021bias} rely on calibration methods. Despite encouraging results, these approaches do not solve the source of the problem which is the bias incurred by the embeddings used.

\section{Fairness in Face Recognition}

In this section, we first briefly recall the main principles of deep Face Recognition and introduce some notations. The interested reader may consult \cite{survey_FR} or \cite{survey_FR2} for a detailed exposition. Then, we present the fairness metrics we adopt and argue of their relevance in our framework.

\subsection{Overview of Face Recognition}\label{subsec:FR_survey}

{\bf Framework.} A typical FR dataset consists of face images of individuals from which we wish to predict the identities. Assuming that the images are of size $h \times w$ and that there are $K$ identities among the images, this can be modeled by i.i.d. realizations of a random variable $(X,y) \in \mathbb{R}^{h \times w \times c} \times \{1, \ldots, K \}$, where $c$ corresponds to the color channel dimension. In the following, we denote by $\mathbb{P}$ the corresponding probability law.

{\bf Objective.} The usual goal of FR is to learn an encoder function $f_\theta: \mathbb{R}^{h \times w \times c} \rightarrow \mathbb{R}^d$ that embeds the images in a way to bring same identities closer together. The resulting latent representation $Z := f_\theta(X)$ is the {\it face embedding} of $X$. Since the advent of deep learning, the encoder is a deep Convolutional Neural Network (CNN) whose parameters $\theta$ are learned on a huge FR dataset $(\vect{x}_i, y_i)_{1 \leq i \leq N}$ made of $N$ i.i.d. realizations of the random variables $(X,y)$. There are generally two FR use cases: {\it identification}, which consists in finding the specific identity of a probe face among several previously enrolled identities, and {\it verification} (which we focus on throughout this paper), which aims at deciding whether two face images correspond to the same identity or not. To do so, the closeness between two embeddings is usually quantified with the cosine similarity measure $s(\vect{z}_i, \vect{z}_j) := \vect{z}_i^\intercal \vect{z}_j / ( || \vect{z}_i || \cdot || \vect{z}_j ||)$, where $|| \cdot ||$ stands for the usual Euclidean norm (the Euclidean metric $|| \vect{z}_i - \vect{z}_j ||$ is also used in some early works {\it e.g.} \cite{facenet}). Therefore, an operating point $t \in [-1, 1]$ (threshold of acceptance) has to be chosen to classify a pair $(\vect{z}_i, \vect{z}_j)$ as {\it genuine} (same identity) if $s \geq t$ and {\it impostor} (distinct identities) otherwise.

{\bf Training.} For the training phase only, a fully-connected layer is added on top of the deep embeddings so that the output is a $K$-dimensional vector, predicting the identity of each image within the training set. The full model (CNN + fully-connected layer) is trained as an identity classification task. Until 2018, most of the popular FR loss functions were of the form:
\begin{equation} \label{eq:standard_loss}
\mathcal{L} = - \frac{1}{n} \sum\limits_{i=1}^n \log \left( \frac{e^{ \kappa \vect{\mu}_{y_i}^\intercal \vect{z}_i}}{\sum_{k=1}^K e^{ \kappa \vect{\mu}_{k}^\intercal \vect{z}_i}} \right),
\end{equation} 
where the $\vect{\mu}_k$'s are the fully-connected layer's parameters, $\kappa > 0$ is the inverse temperature of the softmax function used in brackets and $n$ is the batch size. Early works \cite{deepface,deepID} took $\kappa = 1$ and used a bias term in the fully-connected layer but \cite{normface} showed that the bias term degrades the performance of the model. It was thus quickly discarded in later works. 
Since the canonical similarity measure at the test stage is the cosine similarity, the decision rule only depends on the angle between two embeddings, whereas it could depend on the norms of $\vect{\mu}_k$ and $\vect{z}_i$ during training. This has led \cite{normface} and \cite{vmf_deep_learning} to add a normalization step during training and take $\vect{\mu}_k, \vect{z}_i \in \mathbb{S}^{d-1}:=\{ z \in \mathbb{R}^d: ||z|| = 1\}$ as well as introducing the re-scaling parameter $\kappa$ in Eq.~\ref{eq:standard_loss}: these ideas significantly improved upon former models and are now widely adopted. The hypersphere $\mathbb{S}^{d-1}$ to which the embeddings belong is commonly called face hypersphere.
Denoting by $\theta_{i}$ the angle between $\vect{\mu}_{y_i}$ and $\vect{z}_i$, the major advance over the loss of Eq.~\ref{eq:standard_loss} (with normalization of $\vect{\mu}_k, \vect{z}_i$) in recent years was to consider large-margin losses which replace $\vect{\mu}_{{y_i}}^\intercal \vect{z}_i = \cos(\theta_{i})$ by a function that reduces intra-class angle variations, such as the $\cos(m \theta_{i})$ of \cite{sphereface} or the $\cos(\theta_{i})-m$ of \cite{cosface}. The most efficient choice is $\cos(\theta_{i} + m)$ and is due to \cite{arcface} who called their model ArcFace, on which we build our methodology. A fine training should result in the alignment of each embedding $\vect{z}_i$ with the vector $\vect{\mu}_{y_i}$. The aim is to bring together embeddings with the same identity. Indeed, during the test phase, the learned algorithm will have to decide whether two face images are related to the same, potentially unseen, individual (one refers to an {\it open set} framework).

{\bf Evaluation metrics.} Let $(X_1, y_1)$ and $(X_2,y_2)$ be two independent random variables with law $\mathbb{P}$. We distinguish between the False Acceptance and False Rejection Rates, respectively defined by 
\begin{equation*}
\begin{array}{cc}
 \mathrm{FAR}(t) \! \! \! \! &:= \mathbb{P}( s(Z_1,Z_2) \geq t \ | \ y_1 \neq y_2) \\
 \mathrm{FRR}(t) \! \! \! \! &:= \mathbb{P}( s(Z_1,Z_2) < t \ | \ y_1 = y_2)
\end{array}    
\end{equation*}
These quantities are crucial to evaluate a given algorithm in our context: Face Recognition is intrinsically linked to biometric applications, where the usual accuracy evaluation metric is not sufficient to assess the quality of a learned decision rule. For instance, security automation in an airport requires a very low FAR while keeping a reasonable FRR to ensure a pleasant user experience. As a result, the most widely used metric consists in first fixing a threshold $t$ so that the $\mathrm{FAR}$ is equal to a pre-defined value $\alpha \in [0,1]$, and then computing the $\mathrm{FRR}$ at this threshold. We use the {\it canonical FR notation} to denote the resulting quantity:
\begin{equation*} 
\mathrm{FRR}@(\mathrm{FAR}=\alpha) \, := \, \mathrm{FRR}(t) \, \,  \text{with}  \, \, \mathrm{FAR}(t) = \alpha. 
\end{equation*}
The $\mathrm{FAR}$ level $\alpha$ determines the operational point of the FR system and corresponds to the security risk one is ready to take. According to the use case, it is typically set to $10^{-i}$ with $i \in \{1, \ldots, 6\}$.

\subsection{Incorporating Fairness}\label{subsec:fairness}
While the $\mathrm{FRR}@\mathrm{FAR}$ metric is the standard choice for measuring the performance of a FR algorithm, it does not take into account its variability among different subgroups of the population. In order to assess and correct for potential discriminatory biases, the practitioner must rely on suitable fairness metrics. 

{\bf Framework.} In order to incorporate fairness with respect to a given discrete sensitive attribute that can take $A > 1$ different values, we enrich our previous model and consider a random variable $(X,y,a)$ where $a \in \{0, 1, \ldots, A-1 \}$. With a slight abuse of notations, we still denote by $\mathbb{P}$ the corresponding probability law and, for every fixed value $a$, we can further define
\begin{equation*}
\begin{array}{cc}
 \mathrm{FAR}_a(t) \! \! \! \! &:= \mathbb{P}( s(Z_1,Z_2) \geq t \ | \ y_1 \neq y_2, \ a_1=a_2=a) \\
 \mathrm{FRR}_a(t) \! \! \! \! &:= \mathbb{P}( s(Z_1,Z_2) < t \ | \ y_1 = y_2, \ a_1=a_2=a).
\end{array}    
\end{equation*}
In our case, we focus on gender bias so we take $A=2$ with the convention that $a = 0$ stands for male, $a = 1$ for female. 

{\bf Existing fairness metrics.} Before specifying our choice for the fairness metric used here, let us review some existing ones \cite{survey_fairness_ML} that derive from fairness in the context of binary classification (in FR, one classifies pairs in two groups: genuines or impostors). 
The {\it Demographic Parity} criterion requires the prediction to be independent of the sensitive attribute, which amounts to equalizing the likelihood of being genuine conditional to $a=0$ and $a=1$. Besides heavily depending on the number and quality of impostors and genuines pairs among subgroups, this criterion does not take into account the $\mathrm{FAR}$s and $\mathrm{FRR}$s, which are instrumental in FR as previously mentioned. An attempt to incorporate those criteria could be to compare the intra-group performances: $\mathrm{FRR}_0 @ (\mathrm{FAR}_0 = \alpha)$ v.s. $\mathrm{FRR}_1 @ (\mathrm{FAR}_1 = \alpha)$. However, the operational points $t_0$ and $t_1$ satisfying $\mathrm{FAR_0(t_0)} = \alpha$ and $\mathrm{FAR_1(t_1)} = \alpha$ generically differ as pointed out by \cite{issues_race_bias}. To fairly assess the equity of an algorithm, one needs to compare intra-groups $\mathrm{FAR}$s and $\mathrm{FRR}$s at the same threshold. Two such criteria exist in the fairness literature: the {\it Equal Opportunity} fairness criterion which requires $\mathrm{FRR}_0(t) = \mathrm{FRR}_1(t)$ and the {\it Equalized Odds} criterion which additionally requires $\mathrm{FAR}_0(t) = \mathrm{FAR}_1(t)$. Nevertheless, working at an arbitrary threshold $t$ does not really make sense since, as previously mentioned, FR systems typically choose an operational point achieving a predefined $\mathrm{FAR}$ level so as to limit security breaches. This is why most current papers consider a fixed operational point $t$ such that the global population False Acceptance Rate equals a fixed value $\alpha$. For instance, \cite{dhar2021pass} computes
\begin{equation} 
| \mathrm{FRR}_1(t) - \mathrm{FRR}_0(t) | \quad \text{with} \, \, \mathrm{FAR}(t) = \alpha. 
\label{eq:BiasDelta}
\end{equation}
However, we think that the choice of a threshold achieving a global $\mathrm{FAR}$ is not entirely relevant for it depends on the relative proportions of females and males of the considered dataset together with the relative proportion of intra-group impostors. For instance, at fixed images quality, if females represent a small proportion of the evaluation dataset, the threshold $t$ of Eq. \ref{eq:BiasDelta} is close to the male threshold $t_0$ satisfying $\mathrm{FAR}_0(t_0) = \alpha$ and away from the female threshold $t_1$ satisfying $\mathrm{FAR}_1(t_1)=\alpha$. Such a variability among datasets could lead to incorrect conclusions.

{\bf New fairness metrics. }In this paper, we go one step further and work at a threshold achieving $\max_a \mathrm{FAR}_a=\alpha$ instead of $\mathrm{FAR}=\alpha$. This alleviates the previous proportion dependence. Besides, this allows to monitor the risk one is willing to take among each subgroup: for a pre-definite rate $\alpha$ deemed acceptable, one typically would like to compare the performance among subgroups for a threshold where {\it each} subgroup satisfies $\mathrm{FAR}_a \leq \alpha$. Our two resulting metrics are thus:
\begin{align} \label{eq:fairness_metric1}
\mathrm{BFRR}(\alpha) := \frac{\max_{a \in \{0,1\}} \mathrm{FRR}_a(t)}{\min_{a \in \{0,1\}} \mathrm{FRR}_a(t)} 
\end{align}
and
\begin{equation} \label{eq:fairness_metric2}
\mathrm{BFAR}(\alpha) := \frac{\max_{a \in \{0,1\}} \mathrm{FAR}_a(t)}{\min_{a \in \{0,1\}} \mathrm{FAR}_a(t)},
\end{equation} 
where $t$ is taken such that $\max_{a \in \{0, 1\}} \mathrm{FAR}_a(t) = \alpha $.

One can read the above acronyms ``Bias in $\mathrm{FRR}$/$\mathrm{FAR}$". In addition to being more security demanding than previous metrics, $\mathrm{BFRR}$ and $\mathrm{BFAR}$ are more amenable to interpretation: the ratios of $\mathrm{FRR}$s or $\mathrm{FAR}$s correspond to the number of times the algorithm makes more mistakes on the discriminated subgroup. Those metrics generalize well for more than $2$ distinct values of the sensitive attribute.

\section{Geometric Mitigation of Biases}\label{sec:vMF for eth mod}

Contrary to a common thinking about the origin of bias, training a FR model on a balanced training set (i.e. with as much female identities/images than male identities/images) is not enough to mitigate gender bias in FR \cite{gender_balanced_data}. It is therefore necessary to intervene by designing a model to counteract the gender bias.


\subsection{A Geometrical Embedding View on Fairness}\label{subsec:geometrical_causes_bias}

\begin{figure}[ht!]
    \centering
    \includegraphics[scale=0.4]{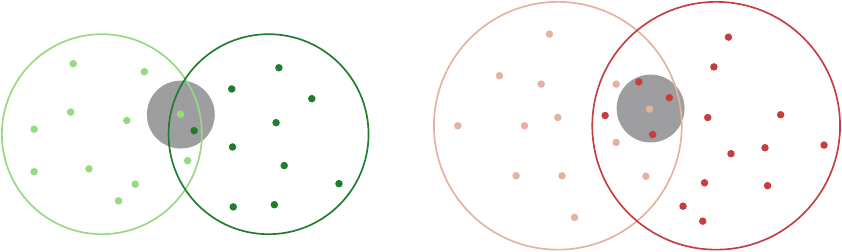}
    \caption{Illustration of the geometric nature of bias. Each point is the embedding of an image. In green: two male identities. In red: two female identities. The overlapping region between two identities is higher for females than for males. The grey circles are the acceptance zones, centered around an embedding of reference, associated to a constant threshold $t$ of acceptance.}
    \label{fig:schema_biais}
    \vspace{-0.3cm}
\end{figure}

In fact, impostor scores (cosine similarities of impostor pairs) are higher for females than for males while genuine scores are lower for females than for males \cite{nist_report_2019, too_bias_or_not}. This puts females at a disadvantage compared to males in terms of both $\mathrm{FAR}$ and $\mathrm{FRR}$. Typically, this is due to {\it(i)} a smaller repulsion between female identities and/or {\it(ii)} a greater intra-class variance (spread of embeddings of each identity) for female identities, as illustrated in \autoref{fig:schema_biais}. Thus, we present in the following a statistical model which enables to set the intra-class variance for each identity on the face hypersphere.

\subsection{von-Mises Fisher Mixture Model}\label{subsec:vMF_Loss}


In order to mitigate the gender bias of deep FR systems, we set a statistical model on the latent representations of images. Recall that we assumed that each individual of a FR dataset is an i.i.d. realization of a random variable $(X,y,a)$, where $X$ is the image, $y$ the identity and $a$ the gender attribute. Also, recall that, both at the training and the testing stages, the embeddings are normalized on the hypersphere, meaning that $Z=f_\theta(X) \in \mathbb{S}^{d-1}$. As previously mentioned, a fine learning should result in an alignment of the embeddings $\{\vect{z}_i\}$ of a same identity $y_i$ around their associated centroid $\vect{\mu}_{y_i} \in \mathbb{S}^{d-1}$. It is therefore reasonable to assume that the embeddings of a same identity are i.i.d. realizations of a radial distribution of gaussian-type on the hypersphere, centered at $\vect{\mu}_{y_i}$. A natural choice is thus to take the so-called von-Mises Fisher (vMF) distribution which is nothing but the law of a gaussian conditioned to live in the hypersphere. Before turning to the formal definition of the statistical model we put on the hypersphere, let us give the definition of this vMF distribution. 

\begin{figure*}[ht!]
    \centering
    \includegraphics[scale=0.45]{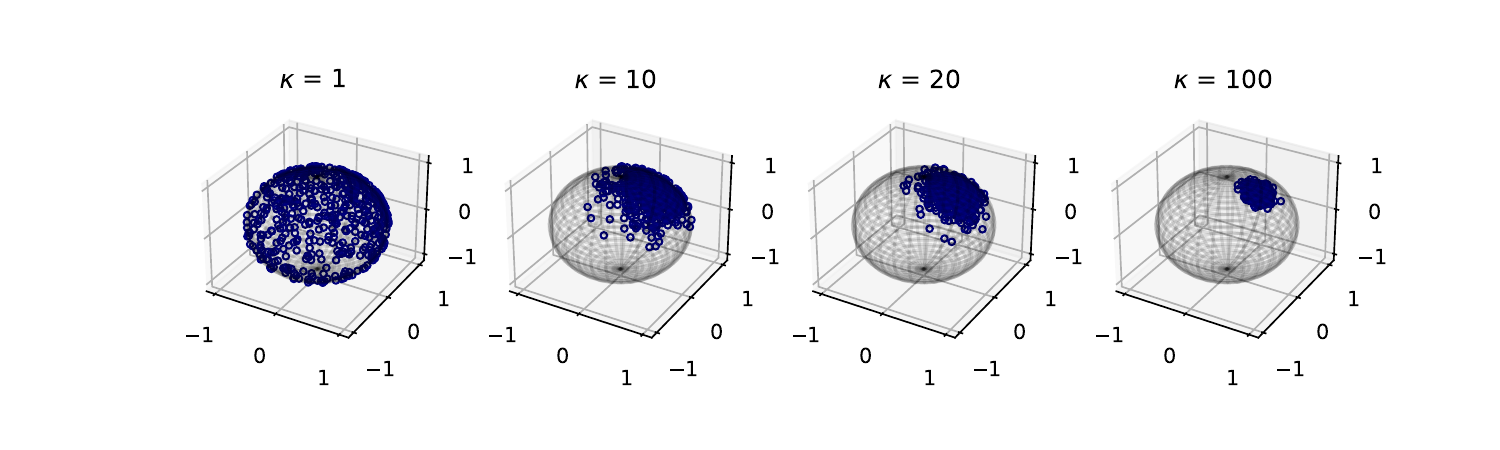}
    \caption{$500$ samples from the vMF distribution in dimension $3$ with parameters $\vect{\mu} =$ [$0.5$, $0$, $\sqrt{0.75}$] and $\kappa > 0$.}
    \label{fig:vMF_kappa_influence}
\end{figure*}

{\bf The von Mises-Fisher distribution.} The vMF distribution in dimension $d$ with mean direction $\vect{\mu} \in \mathbb{S}^{d-1}$ and concentration parameter $\kappa > 0$ is a probability measure defined on the hypersphere $\mathbb{S}^{d-1}$ by the following density:
\[ V_d( \vect{z} ; \vect{\mu}, \kappa) := C_d(\kappa) e^{\displaystyle \kappa  \vect{\mu}^\intercal \vect{z}},  \]
with $C_d(\kappa) = \kappa^{\frac{d}{2} - 1} / ((2 \pi)^{\frac{d}{2}} I_{\frac{d}{2} - 1}(\kappa) )$. $I_\nu$ stands for the modified Bessel function of the first kind at order $\nu$, whose logarithm can be computed with high precision (see supplementary material \ref{app:vMF_constants}). The vMF distribution corresponds to a gaussian distribution in dimension $d$ with mean $\vect{\mu}$ and
covariance matrix $(1/\kappa)I_d$, conditioned to live on $\mathbb{S}^{d-1}$.
\autoref{fig:vMF_kappa_influence} illustrates the influence of the concentration parameter~$\kappa$ on the vMF distribution.

\begin{figure}
\vspace{-0.3cm}
    \centering
    \includegraphics[scale=0.3]{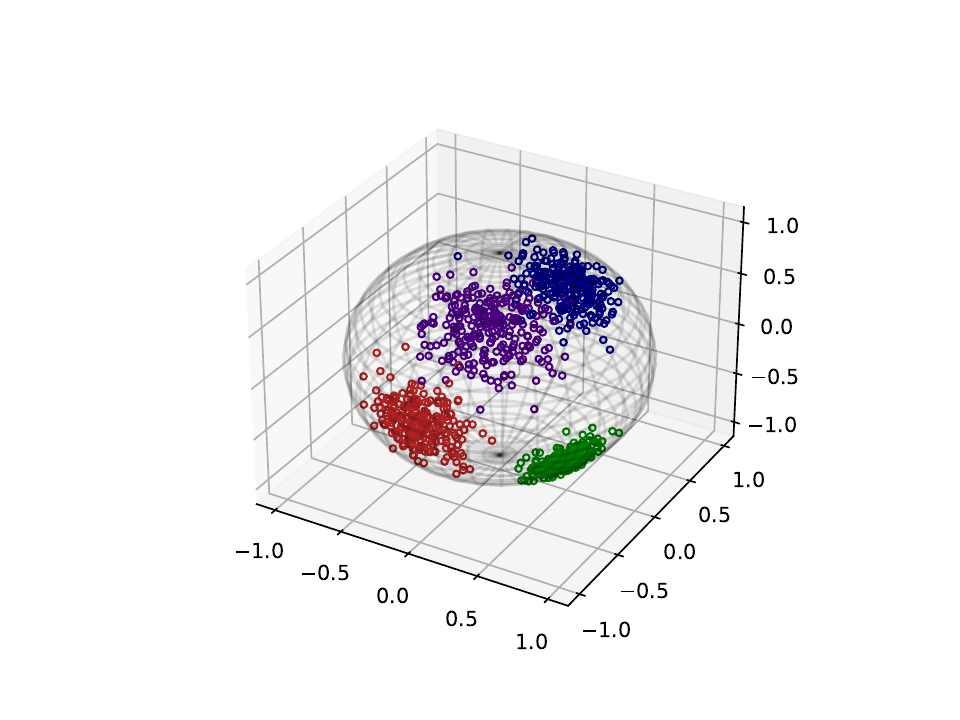}
    \caption{Illustration of a
vMF mixture model.}
    \vspace{-0.5cm}
    \label{fig:vMF_mixture_model}
\end{figure}

{\bf Mixture model.} Since the vMF distribution seems to reflect well the distribution of the embeddings of $1$ identity around their centroid, we extend the model to include all the $K$ identities from the training set by considering a mixture model where each component $k$ ($1 \leq k \leq K$) is equiprobable and follows a vMF distribution $V_d( \vect{z} ; \vect{\mu}_k, \kappa_k)$. \autoref{fig:vMF_mixture_model} provides an illustration of the mixture model.

{\bf Maximum likelihood.} Let $N\geq 1$ and $(\vect{x}_i, y_i, a_i)_{1 \leq i \leq N}$ be i.i.d. realizations of $(X,y,a)$. Under the previous vMF mixture model assumption, the probability $p_{ij}$ that a face embedding $\vect{z}_i = f_\theta(\vect{x}_i)$ belongs to identity $j$ is given by
\[ p_{ij} = \frac{V_d(\vect{z}_i|\vect{\mu}_j, \kappa_j)}{\sum_{k=1}^K V_d(\vect{z}_i|\vect{\mu}_k, \kappa_k)}= \frac{  C_d(\kappa_j) \ e^{\displaystyle \kappa_j \ \vect{\mu}_j^\intercal \vect{z}_i}}{\sum_{k=1}^K  C_d(\kappa_k) \ e^{\displaystyle \kappa_k \ \vect{\mu}_k^\intercal \vect{z}_i}}. \]
Therefore, the negative log-likelihood of the model is
\begin{equation} \label{eq:NLL-vMF}
   \mathrm{NLL} =  -\frac{1}{N} \sum_{i=1}^N \log \left[ \frac{ C_d(\kappa_{y_i}) \ e^{\displaystyle \kappa_{y_i} \ \vect{\mu}_{y_i}^\intercal \vect{z}_i}}{\sum_{k=1}^K  C_d(\kappa_k) \ e^{\displaystyle \kappa_k \ \vect{\mu}_k^\intercal \vect{z}_i}} \right].
\end{equation}  

In that case, the above NLL is in fact the vMF loss function, firstly introduced in
the context of FR by \cite{vmf_deep_learning}  who took a unique hyperparameter value $\kappa$. In this situation, the vMF loss reduces to the classical loss of Eq.~\ref{eq:standard_loss} (when $\vect{z}_i$ and $\vect{\mu}_k$ are normalized). This makes the vMF loss a natural generalization of popular FR loss functions, before the advent of large-margin losses. \cite{adaptive_margin_loss} introduce a similar loss with $2$ distinct $\kappa$ values but they do not take into account the normalization constant $C_d(\kappa)$. \cite{vmf_classification} use the vMF loss with unique concentration parameter $\kappa$ for image classification and retrieval but the centroids $\vect{\mu_k}$ are not learned by gradient descent but rather by an approximate maximum likelihood estimation.


{\bf Training an Ethical Module with the vMF-loss.}
In order to correct for the gender bias contained within the learned latent representation, we train a shallow MultiLayer Perceptron (MLP) which is designed to give more representation power to females. To do so, we slightly modify Eq.~\ref{eq:NLL-vMF} by replacing the concentration parameter $\kappa_k$ of each identity $k$ by a concentration parameter that only depends on the gender $a_k \in \{0,1\}$ of the identity $k$. In other words, we replace $\kappa_k$ by $\kappa_{a_k}$ and we end up with only $2$ concentration parameters ($\kappa_0, \kappa_1 >0$) that we take as hyperparameters. To sum up, we train the MLP with the following {\it Fair von Mises-Fisher loss} (FvMF), on batches of size $n$:
\begin{equation}\label{eq:FvMF_loss}
\mathcal{L}_{\text{FvMF}} = -\frac{1}{n} \sum\limits_{i=1}^n \log \left( \frac{C_d(\kappa_{a_{y_i}}) e^{\displaystyle \kappa_{a_{y_i}}  \vect{\mu}_{y_i}^\intercal \vect{z}_i}}{ \sum_{k=1}^K  C_d(\kappa_{a_k}) e^{\displaystyle \kappa_{a_k} \vect{\mu}_k^\intercal \vect{z}_i} } \right).
\end{equation}
Notice that there are two ways of minimizing $\mathcal{L}_{\text{FvMF}}$: either by aligning normalized face embeddings $\vect{z}_i$ with associated ground-truth $\vect{\mu}_{y_i}$ (the intra-class variance is characterized by $\kappa_{a_{y_i}}$) or by pushing back wrong $\vect{\mu}_k$ (with $k \neq y_i$) from $\vect{z}_i$ (the repulsion strength is related to $\kappa_{a_k}$). This brings us back to the two geometric causes for bias in FR, presented in section \ref{subsec:geometrical_causes_bias}.
However, those two phenomena are in competition during the loss minimization, especially with two distinct values of concentration parameter, which makes it difficult to predict the optimal values of $\kappa_0$ and $\kappa_1$.

\section{Numerical Experiments} \label{sec:expe}

{\bf Pre-trained models.} We use the trained model ArcFace\footnote{\url{https://github.com/deepinsight/insightface/tree/master/recognition/arcface_torch}.} whose CNN architecture is a ResNet100 \cite{resnet100_forFR}. As emphasized before, it achieves state-of-the-art performances in FR. It has been trained on the MS1M-RetinaFace dataset (also called MS1MV3), introduced by \cite{ms1m_retinaface} in the ICCV 2019 Lightweight Face Recognition Challenge. MS1MV3 is a cleaned version of the MS-Celeb1M dataset \cite{ms-celeb-1m}; all its face images have been pre-processed by the Retina-Face detector \cite{retinaface_detector} and are of size $112\times112$ pixels. It  contains $5.1$M images of $93$k identities. We also consider other pre-trained models\footnote{\label{footnote:faceX_zoo}\url{https://github.com/JDAI-CV/FaceX-Zoo/blob/main/training_mode/README.md}.} (AdaCos~\cite{adacos}, CosFace~\cite{cosface}, CurricularFace~\cite{curricularface}) whose backbone is a MobileFaceNet~\cite{mobilefacenets}, trained on the MS-Celeb-1M-v1c-r dataset\footnote{See footnote \ref{footnote:faceX_zoo}.}. This dataset is another cleaned version of the MS-Celeb1M dataset and it contains $3.28$M images of $73$k identities. The images are also pre-processed by the Retina-Face detector and are of size $112\times112$ pixels.

{\bf Gender labels.} For a fair comparison, we train our Ethical Module on the training set used to train the pre-trained models (MS1MV3 for ArcFace, MS-Celeb-1M-v1c-r for the models with MobileFaceNet backbone). However, ground-truth gender labels for MS1MV3/MS-Celeb-1M-v1c-r are not available. As the training of our Ethical Module needs the gender label of each face image within the training set, we use a private gender classifier to get those gender labels. Current gender classifiers achieve around $95$\% prediction accuracy on standard evaluation datasets and are widely used in FR to get gender annotations \cite{measure_privacy, gac}. Since some images from the same identity might be assigned different gender predictions, it is common practice to use a majority vote to decide the correct gender for each identity. We follow \cite{gender_balanced_data} and only keep in our training sets the identities for which at least $75$\% of the same-identity face images are assigned the same gender. Doing so, we discard $25$k images and $835$ identities for MS1MV3, $10$k images and $500$ identities for MS-Celeb-1M-v1c-r.

\begin{figure*}[ht!]
\vspace{-0.2cm}
\hspace{-1cm}
    \includegraphics[scale=0.33]{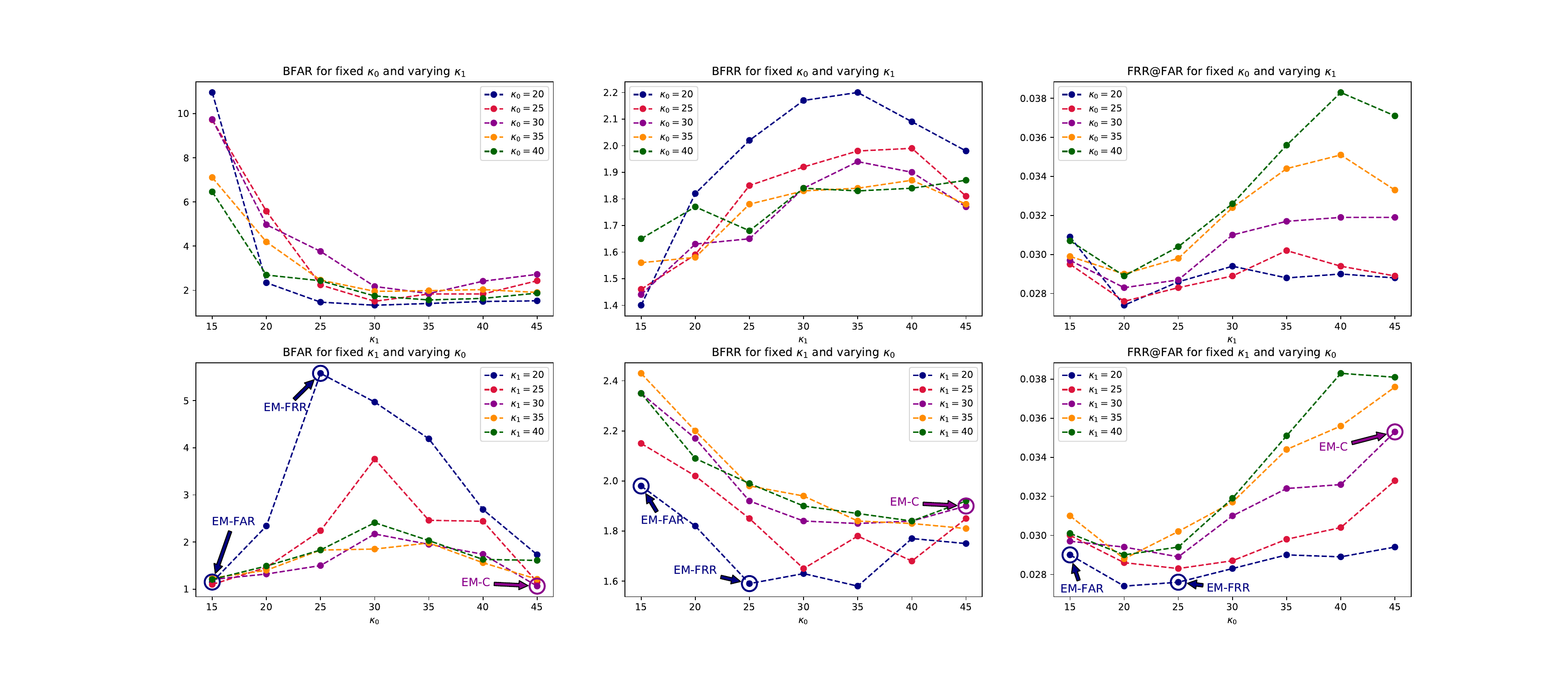}
    \vspace{-0.9cm}
    \caption{Fairness and evaluation metrics on IJB-C for the Ethical Module when one of the two hyperparameters is fixed. The FAR level defining the threshold $t$ is set to $10^{-3}$; the pre-trained model is ArcFace with a ResNet100 backbone. $\mathrm{FRR}@\mathrm{FAR}$ is expressed as a percentage (\%). The three versions of the Ethical Module presented in \autoref{tab:kappa_choice} are annotated with circles.}
    \label{fig:hyperparam_trends}
    \vspace{-0.1in}
\end{figure*}

{\bf Ethical Module.} The face embeddings output by the pre-trained models are of dimension $512$. Thus, the MLP within our Ethical Module has an input layer of $512$ units. To emphasize the fact that our gender bias mitigation solution is much less costly than current solutions such as \cite{mitigating_bias_RL} and \cite{dhar2021pass}, in terms of both training time and computation power (see supplementary material~\ref{app:numerical_stability}), we choose a shallow MLP of size ($512$,~$1024$,~$512$) with a ReLU activation after the first layer, the output dimension being the same than for the pre-trained models. This MLP is trained with the fair version $\mathcal{L}_{\text{FvMF}}$ of the vMF loss, introduced in Eq. \ref{eq:FvMF_loss}. For each experiment, we train the Ethical Module during $50$ epochs with the Adam optimizer \cite{adam}. The batch size is set to $1024$ and the learning rate to $0.01$. The training is efficient as we first compute the face embeddings of the pre-trained models (on MS1MV3 for ArcFace, on MS-Celeb-1M-v1c-r for the models with MobileFaceNet backbone), store them, and then train a shallow MLP on those embeddings. Using one single GPU (NVIDIA RTX 3090), the computation of the embeddings takes $4$ hours and each training takes $8$ hours.

{\bf Reproducibility.} The code used for the experiments can be found at \url{https://github.com/JRConti/EthicalModule_vMF}.

\subsection{Grid-Search on IJB-C}\label{subsec:grid_search_ijbc}

In order to select relevant pairs of gender-hyperparameters $(\kappa_0, \kappa_1)$, we perform a grid-search and keep track of the canonical performance metric $\mathrm{FRR}@(\mathrm{FAR}=10^{-3})$ together with our two fairness metrics $\mathrm{BFRR}(10^{-3})$ and $\mathrm{BFAR}(10^{-3})$ introduced in Eq.~\ref{eq:fairness_metric1} and \ref{eq:fairness_metric2}. To obtain reliable results, we need to compute the latter metrics on a sufficiently large FR dataset containing gender labels. We choose IJB-C \cite{maze2018iarpa}, which contains about 3,5k identities for a total number of about $31$k images and $117$k unconstrained video frames. The 1:1 verification protocol\footnote{\url{https://github.com/deepinsight/insightface/tree/master/recognition/_evaluation_/ijb}.}  is performed on $19$k genuine pairs and $15$M impostor pairs. We choose ArcFace ResNet100 as the pre-trained model for this experiment. The results are displayed in \autoref{fig:hyperparam_trends}.

Several interesting trends emerge from \autoref{fig:hyperparam_trends}, suggesting an underlying regularity of the model with respect to the hyperparameters' space. More precisely, when $\kappa_0$ is fixed and $\kappa_1$ increases, $\mathrm{BFAR}$ tends to decrease, $\mathrm{BFRR}$ first increases and then decreases and $\mathrm{FRR} @ \mathrm{FAR}$ tends to increase. When $\kappa_1$ is fixed and $\kappa_0$ increases, $\mathrm{BFAR}$ first increases and then decreases, $\mathrm{BFRR}$ tends to decrease and $\mathrm{FRR} @ \mathrm{FAR}$ increases. In the supplementary material \ref{app:explain_trends_grid_search}, we give some explanations of the trends in \autoref{fig:hyperparam_trends}. Note that BFAR and BFRR have opposite behaviors, which reveals a trade-off between both fairness metrics.

Many  ($\kappa_0$, $\kappa_1$) pairs could be considered as relevant and instead of defining an objective criterion, we select three of them in order to illustrate the trade-offs one needs to make between fairness metrics and pure performance. The selection is made based on \autoref{fig:hyperparam_trends}. We provide in \autoref{tab:kappa_choice} the ($\kappa_0$, $\kappa_1$) pairs which optimize each pair of the  considered metrics and give them a name for what follows. The three versions of the Ethical Module presented in \autoref{tab:kappa_choice} are robust to a change of FAR level, when performing the grid-search, as illustrated in the supplementary material~\ref{app:robustness_kappas}.


\begin{table}[h!]
\caption{Hyperparameters selected to optimize each pair of metrics. We give a name to each of the ($\kappa_0$, $\kappa_1$) pairs. EM stands for Ethical Module.}\label{tab:kappa_choice}
\vskip 0.1in
\begin{center}
\begin{small}
\begin{sc}
\begin{tabular}{lccccr}
\toprule
Name & $\mathrm{BFRR}$ & $\mathrm{BFAR}$ & $\mathrm{FRR}@\mathrm{FAR}$ & $\kappa_0$ & $\kappa_1$ \\
\midrule 
EM-FAR & $\times$ & $\surd$ & $\surd$ & 15 & 20\\
EM-FRR & $\surd$ & $\times$ & $\surd$ & 25 & 20\\
EM-C & $\surd$ & $\surd$ & $\times$ & 45 & 30\\
\bottomrule
\end{tabular}
\end{sc}
\end{small}
\end{center}
\vskip -0.2in
\end{table}


\begin{table*}[h!]
\caption{Evaluation on LFW for ArcFace with ResNet50 backbone. $\mathrm{FRR}@\mathrm{FAR}$ is expressed as a percentage (\%). $\mathbf{Bold}$=Best, \underline{Underlined}=Second best.}
\vskip 0.1in
\begin{center}
\begin{small}
\begin{sc}
\begin{tabular}{ c | ccc | ccc}
 $\mathrm{FAR}$ level:          & \multicolumn{3}{c}{ $10^{-4}$} & \multicolumn{3}{c}{ $10^{-3}$} \\ 
 \toprule
  Model     & $\mathrm{FRR}@\mathrm{FAR}$ (\%)  & $\mathrm{BFRR}$  & $\mathrm{BFAR}$  & $\mathrm{FRR}@\mathrm{FAR}$ (\%)  & $\mathrm{BFRR}$        & $\mathrm{BFAR}$        \\ \midrule 
                               ArcFace  & $\mathbf{0.078}$  & $10.27$ & $4.72$  & $\underline{0.059}$   & $\underline{4.17}$  & $1.81$ \\ 
                ArcFace + PASS-g & $0.315$ & $\mathbf{4.54}$ & $6.51$ & $0.107$ & $5.22$ & $2.11$\\
                               
   ArcFace + EM-FAR & $0.151$  & $11.22$ & $\mathbf{2.11}$  & $0.072$ & $9.16$ &  $\mathbf{1.19}$ \\
                               ArcFace + EM-FRR & $\underline{0.100}$ & $\underline{5.89}$ & $33.65$ & $\mathbf{0.058}$ & $\mathbf{4.11}$  & $5.24$ \\
                               ArcFace + EM-C & $0.164$  & $9.18$ & $\underline{2.44}$ & $0.081$ & $5.15$  & $\underline{1.20}$  \\ \bottomrule
\end{tabular}
\end{sc}
\end{small}
\end{center}
\vskip -0.1in
\label{table:LFW_1vs1_models}
\end{table*}

\subsection{Fairness Evaluation on LFW}

In this section, we evaluate the three versions of our Ethical Module (EM-FAR, EM-FRR, EM-C) and we compare them to the pre-trained model in terms of fairness and performance. All the models are evaluated on the LFW dataset \cite{LFW_dataset}. The official LFW protocol only considers a few matching pairs among all the possible pairs given the whole LFW dataset. The number of female images is typically not enough to get good estimates of our fairness metrics. To overcome this, we consider all possible same-gender matching pairs among the whole LFW dataset. Doing so, we obtain $9.8$k female genuine pairs, $232$k male genuine pairs, $4.4$M female impostor pairs and $52$M male impostor pairs.

{\bf Baseline.} The current state-of-the-art post-processing method for gender bias mitigation of FR models is achieved by PASS-g \cite{dhar2021pass}. It also consists in transforming the embeddings output by the pre-trained model but it is trained in an adversarial way to classify identities and simultaneously reduce encoding of gender within the new embeddings. Although attempting to output embeddings that are independent from gender seems a good idea, we believe that the gender information contained within the embeddings helps any FR model a lot at the training stage (identity classification), and thus that such a training cannot be achieved without losing too much performance.

We first verify the effectiveness of our Ethical Module using the pre-trained model ArcFace ResNet50. For a fair comparison, we train PASS-g on the same training set than the Ethical Module (MS1MV3 in this case). In \autoref{table:LFW_1vs1_models}, we summarize the different metrics evaluated for the three versions of our Ethical Module on the LFW dataset and compare them with the pre-trained ArcFace and PASS-g baselines, at two $\mathrm{FAR}$ levels. EM-FAR achieves the best $\mathrm{BFAR}$ at both $\mathrm{FAR}$ levels while the best $\mathrm{BFRR}$ is obtained by EM-FRR at $\mathrm{FAR}=10^{-3}$ and by PASS-g at $\mathrm{FAR}=10^{-4}$. At the latter $\mathrm{FAR}$ level, the error rate $\mathrm{FRR}@\mathrm{FAR}$ of PASS-g is slightly more than $4$ times the error rate of the original pre-trained model. Finally, EM-C is the only model which succeeds in reducing both fairness metrics ($\mathrm{BFRR}$ and $\mathrm{BFAR}$) of the pre-trained model at the same time for  $\mathrm{FAR}=10^{-4}$.

In addition, we check the robustness of our method to a change of pre-trained model by considering competitive FR loss functions (AdaCos, CosFace, CurricularFace) with MobileFaceNet backbone. The results are displayed in \autoref{tab:LFW_mobilefacenet}. Additional results with other pre-trained models are available in the supplementary material \ref{app:additional_results}.


\begin{table*}
\caption{Evaluation on LFW for different pre-trained models (AdaCos, CosFace, CurricularFace) with MobileFaceNet backbone. By "Original" we mean no Ethical Module is added to the pre-trained model. $\mathrm{FRR}@\mathrm{FAR}$ is expressed as a percentage~(\%). $\mathbf{Bold}$=Best, \underline{Underlined}=Second best.}
\vskip 0.1in
\begin{center}
\begin{small}
\begin{sc}
\begin{tabular}{ c | ccc | ccc}
  $\mathrm{FAR}$ level:         & \multicolumn{3}{c}{ $10^{-4}$} & \multicolumn{3}{c}{ $10^{-3}$} \\ 
 \toprule
  Model    & $\mathrm{FRR}@\mathrm{FAR}$ (\%)  & $\mathrm{BFRR}$  & $\mathrm{BFAR}$  & $\mathrm{FRR}@\mathrm{FAR}$ (\%)  & $\mathrm{BFRR}$        & $\mathrm{BFAR}$        \\ \midrule \midrule
  AdaCos&&&&&&\\
\midrule                              Original  & $\mathbf{2.97}$ & $\underline{3.64}$ & $3.84$ & $\underline{0.98}$  & $\underline{5.29}$ & $2.23$ \\
  EM-FAR & $4.56$ & $4.42$ & $\mathbf{1.41}$  & $1.33$ & $6.34$ & $\mathbf{1.01}$  \\
                 EM-FRR & $\underline{3.12}$ & $\mathbf{2.71}$ & $8.37$ & $\mathbf{0.91}$ & $\mathbf{4.23}$  & $3.71$ \\
         EM-C & $4.05$ & $4.51$  & $\underline{1.57}$ & $1.26$ & $7.28$ & $\underline{1.08}$  \\ \midrule \midrule
  CosFace&&&&&&\\
\midrule
             Original  & $\mathbf{1.73}$ & $5.89$ & $\underline{2.51}$ & $\mathbf{0.58}$  & $8.18$  & $\underline{1.74}$  \\
 EM-FAR & $3.69$ & $5.76$ & $\mathbf{1.13}$  & $1.05$ & $8.41$ &  $\mathbf{1.02}$ \\
 EM-FRR & $\underline{2.41}$ & $\mathbf{3.03}$ & $9.66$ & $\underline{0.67}$ & $\mathbf{5.09}$  & $4.75$ \\
 EM-C & $2.60$ & $\underline{4.30}$ & $3.69$ & $0.82$ & $\underline{6.81}$ & $1.87$ \\ 
 \midrule \midrule
  CurricularFace&&&&&&\\
\midrule
 Original  & $\mathbf{2.52}$ & $3.67$ & $2.92$ & $\mathbf{0.81}$  & $\underline{4.88}$ & $1.91$  \\
   EM-FAR & $3.86$ & $5.26$ & $\mathbf{1.16}$  & $1.17$ & $6.35$  & $\mathbf{1.10}$  \\
 EM-FRR & $\underline{2.82}$ & $\mathbf{2.58}$ & $9.10$ & $\underline{0.82}$ & $\mathbf{3.89}$  & $4.28$ \\
 EM-C & $3.61$ & $\underline{3.40}$ & $\underline{2.30}$ & $1.02$ & $5.63$ & $\underline{1.27}$  \\ \bottomrule
\end{tabular}
\label{tab:LFW_mobilefacenet}
\end{sc}
\end{small}
\end{center}
\vskip -0.1in
\end{table*}



\section{Conclusion}

In this paper, we introduce a novel method, the {\it Ethical Module}, to mitigate the gender bias of Face Recognition state-of-the-art models. It consists in learning a shallow MLP on top of a frozen pre-trained model, so as to correct the biases that exist in the embedding space. To achieve fairness, we rely on a fair version of the von Mises-Fisher loss that incorporates an hyperparameter per gender, related to the intra-class variance of each gender. Measuring the fairness of Face Recognition systems is a very challenging task and we introduce two new metrics, BFAR and BFRR, that both respond to the need for security and equity. 

Besides being very simple, the resulting methodology is more stable and faster than most current methods of bias mitigation. It both leverages the strong accuracy of pre-trained models while correcting their bias. We illustrate the soundness of our methodology on several pre-trained models, and strongly believe it could also be used to alleviate other types of bias. Our work opens several lines of research: for instance, it would be interesting to extend our ideas to the context of multiclass sensitive attributes and of continuous sensitive attributes such as age. Another idea would be to somehow incorporate our fairness criteria during the training of the Ethical Module. Finally, we think that incorporating large-margin constraints into the loss used to train the Ethical Module would be a promising attempt to go beyond the trade-off between fairness and performance.

\section*{Acknowledgments}
This research was partially supported by the French National Research Agency (ANR), under grant ANR-20-CE23-0028 (LIMPID project).

\bibliography{biblio}
\bibliographystyle{icml2022_env/icml2022}

\newpage
\appendix
\onecolumn

\section{Numerical stability}\label{app:numerical_stability}

\subsection{von Mises-Fisher constants}\label{app:vMF_constants}

Recall the loss defined in Equation~\ref{eq:FvMF_loss}:
\begin{equation*} 
\mathcal{L}_{\text{FvMF}} = -\frac{1}{n} \sum\limits_{i=1}^n \log \left( \frac{C_d(\kappa_{a_{y_i}}) e^{ \kappa_{a_{y_i}}  \vect{\mu}_{y_i}^\intercal \vect{z}_i}}{ \sum_{k=1}^K  C_d(\kappa_{a_k}) e^{ \kappa_{a_k} \vect{\mu}_k^\intercal \vect{z}_i} } \right) \quad \text{with} \quad C_d(\kappa) = \frac{\kappa^{\frac{d}{2} - 1}}{(2 \pi)^{\frac{d}{2}} I_{\frac{d}{2} - 1}(\kappa) }.
\end{equation*}

$I_\nu$ stands for the modified Bessel function of the first kind at order $\nu$, whose logarithm can be computed with high precision using a Python library for arbitrary-precision floating-point arithmetic such as \texttt{mpmath}
\cite{mpmath,vmf_pytorch_implementation}.

\begin{figure}[ht!]
    \centering
    \includegraphics[scale=0.45]{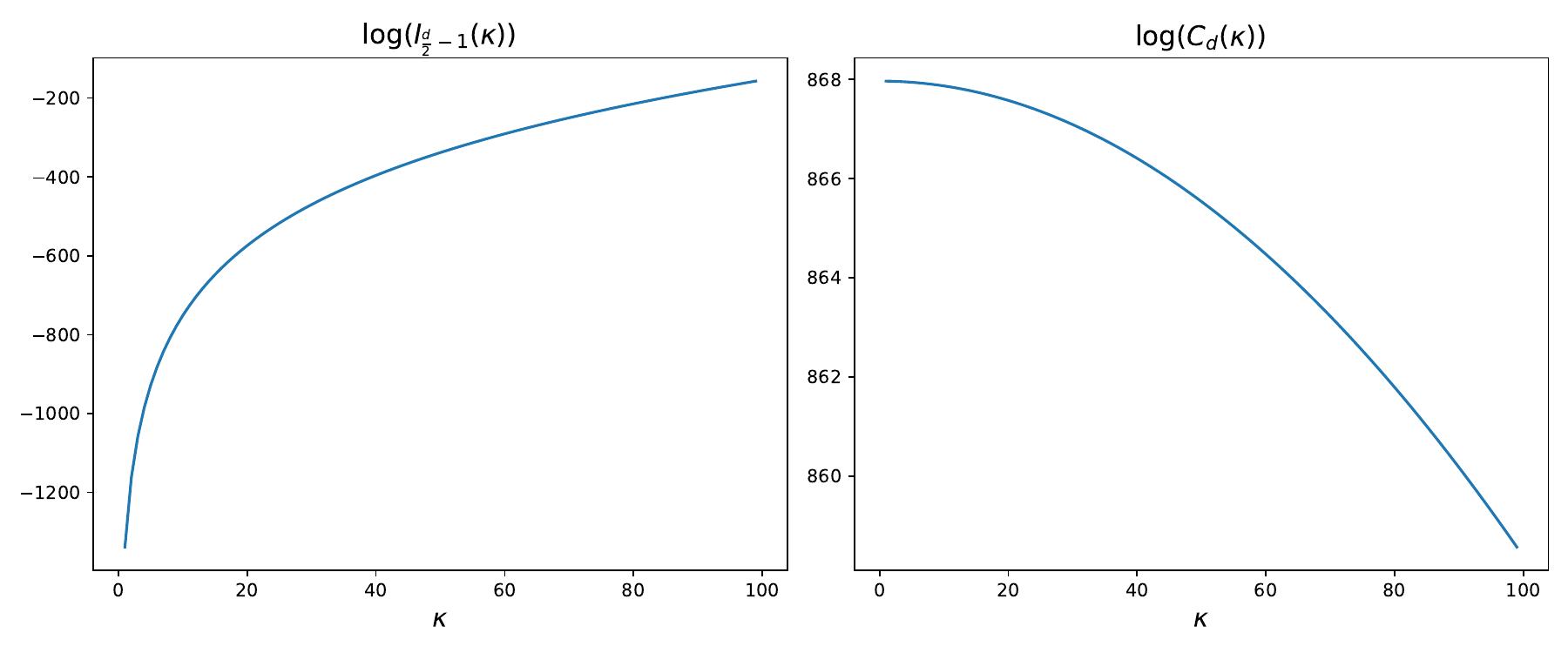}
    \caption{$\log(I_{\frac{d}{2} - 1}(\kappa))$ and $\log(C_d(\kappa))$ as functions of $\kappa$ for $d = 512$.}
    \label{fig:vMF_logpartition}
\end{figure}

Once $\log(I_{\frac{d}{2} - 1}(\kappa))$ is obtained, one is able to compute the logarithm of $C_d(\kappa)$ as:
\begin{equation*}
    \log(C_d(\kappa)) = (\frac{d}{2}-1) \log(\kappa) - \frac{d}{2} \log(2\pi) - \log(I_{\frac{d}{2} - 1}(\kappa)).
\end{equation*}
Figure~\ref{fig:vMF_logpartition} displays $\log(I_{\frac{d}{2} - 1}(\kappa))$ and $\log(C_d(\kappa))$ as functions of $\kappa$ for $d = 512$.

\subsection{Loss stability}

To make use of the numerical stability of the quantity $\log(C_d(\kappa))$, $\mathcal{L}_{\text{FvMF}}$ can be written as:
\begin{equation*} 
\mathcal{L}_{\text{FvMF}} = -\frac{1}{n} \sum\limits_{i=1}^n \log \left( \frac{ e^{\log(C_d(\kappa_{a_{y_i}})) + \kappa_{a_{y_i}}  \vect{\mu}_{y_i}^\intercal \vect{z}_i}}{ \sum_{k=1}^K e^{ \log(C_d(\kappa_{a_k})) + \kappa_{a_k} \vect{\mu}_k^\intercal \vect{z}_i} } \right).
\end{equation*}

Recall the cross-entropy loss $\mathcal{L}_{CE}(\{q_{i,k}\}, \{y_i\})$ with $1\leq i \leq n$ and $1 \leq k \leq K$ defined as:
\begin{equation*}
    \mathcal{L}_{CE}(\{q_{i,k}\}, \{y_i\}) = -\frac{1}{n} \sum\limits_{i=1}^n \log \left( \frac{ e^{q_{i, y_i}} }{ \sum_{k=1}^K e^{q_{i,k}}} \right) 
\end{equation*}

$\mathcal{L}_\text{FvMF}$ can be expressed as the cross-entropy loss:
\begin{equation*}
 \mathcal{L}_\text{FvMF} = \mathcal{L}_{CE} (\{q_{i,k}\}, \{y_i\}) 
\end{equation*}
where the logits $q_{i,k} = \log(C_d(\kappa_{a_k})) + \kappa_{a_k} \vect{\mu}_k^\intercal \vect{z}_i$ satisfy ($\vect{\mu}_k$, $\vect{z}_i \in \mathbb{S}^{d-1}$):
\begin{equation*}
\log(C_d(\kappa_{a_k})) - \kappa_{a_k} \leq q_{i,k} \leq \log(C_d(\kappa_{a_k})) + \kappa_{a_k}
\end{equation*}

\begin{figure}[ht!]
    \centering
    \includegraphics[scale=0.45]{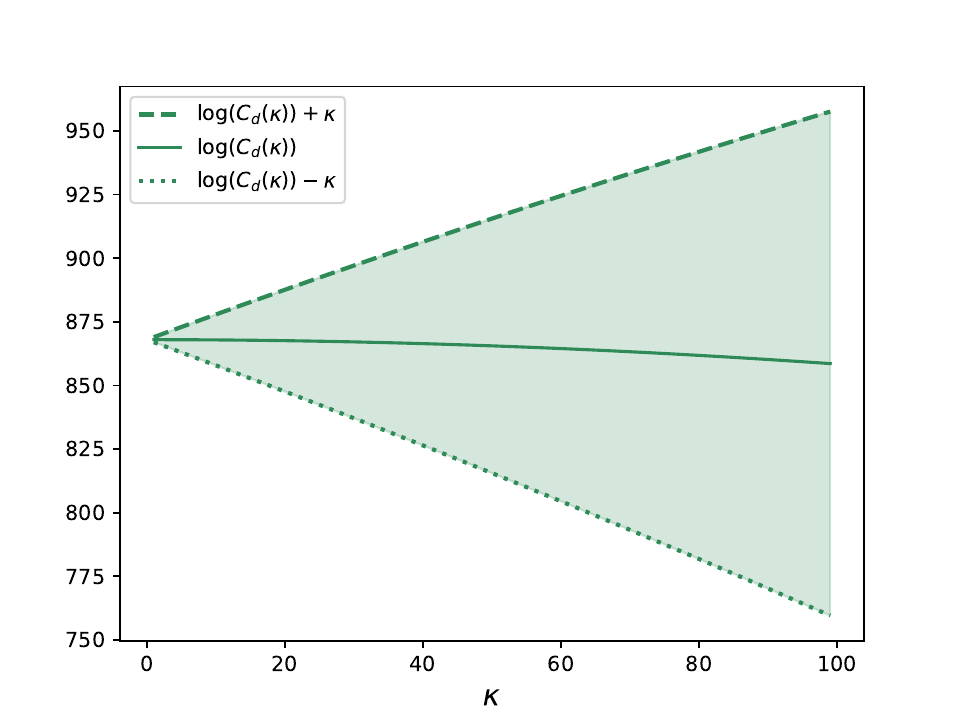}
    \caption{Range of values of the $\mathcal{L}_\text{FvMF}$ loss logits for $d = 512$.}
    \label{fig:vMF_logpartition_bounds}
\end{figure}

Those bounds are displayed in Figure~\ref{fig:vMF_logpartition_bounds}. The fact that $\mathcal{L}_\text{FvMF}$ can be expressed as the cross-entropy loss makes it possible to use the logsoftmax trick and thus further increases its numerical stability. 

\begin{figure}[ht!]
    \centering
    \includegraphics[scale=0.5]{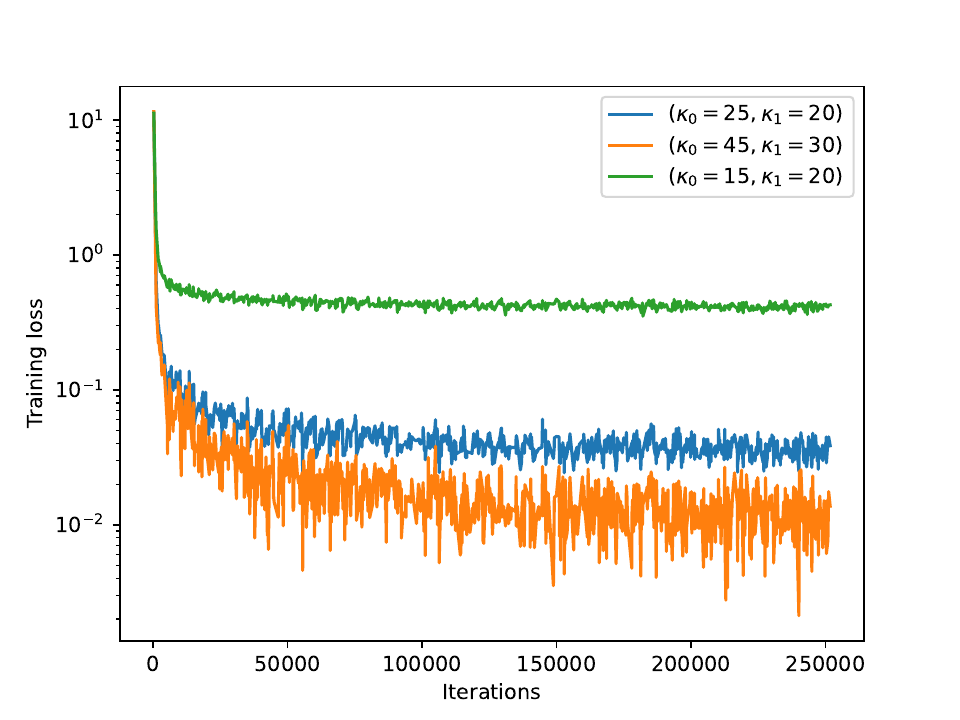}
    \caption{$\mathcal{L}_\text{FvMF}$ training loss for the pre-trained model ArcFace ResNet50.}
    \label{fig:train_loss}
\end{figure}

We provide on Figure~\ref{fig:train_loss} the behavior of our $\mathcal{L}_\text{FvMF}$ training loss, used to train the Ethical Module on top of ArcFace ResNet50.


\section{Grid-search on IJB-C}\label{app:grid_search_2d}

\begin{figure}[ht!]
    \centering
    \subfigure[$\mathrm{FRR}_1(t) / \mathrm{FRR}_0(t)$]{\includegraphics[width=80mm]{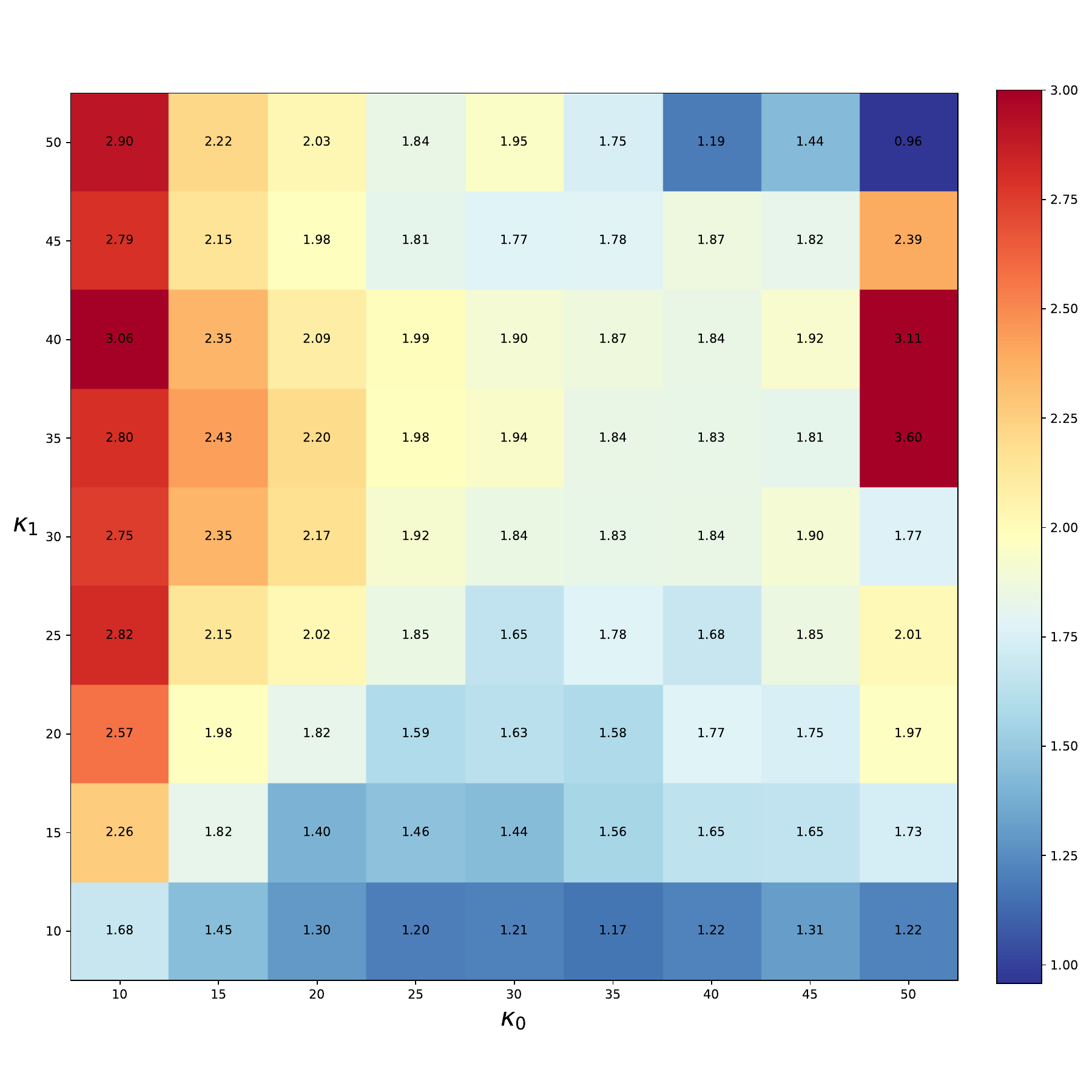}\label{fig:sub1}}
    \subfigure[$\mathrm{FAR}_1(t) / \mathrm{FAR}_0(t)$]{\includegraphics[width=80mm]{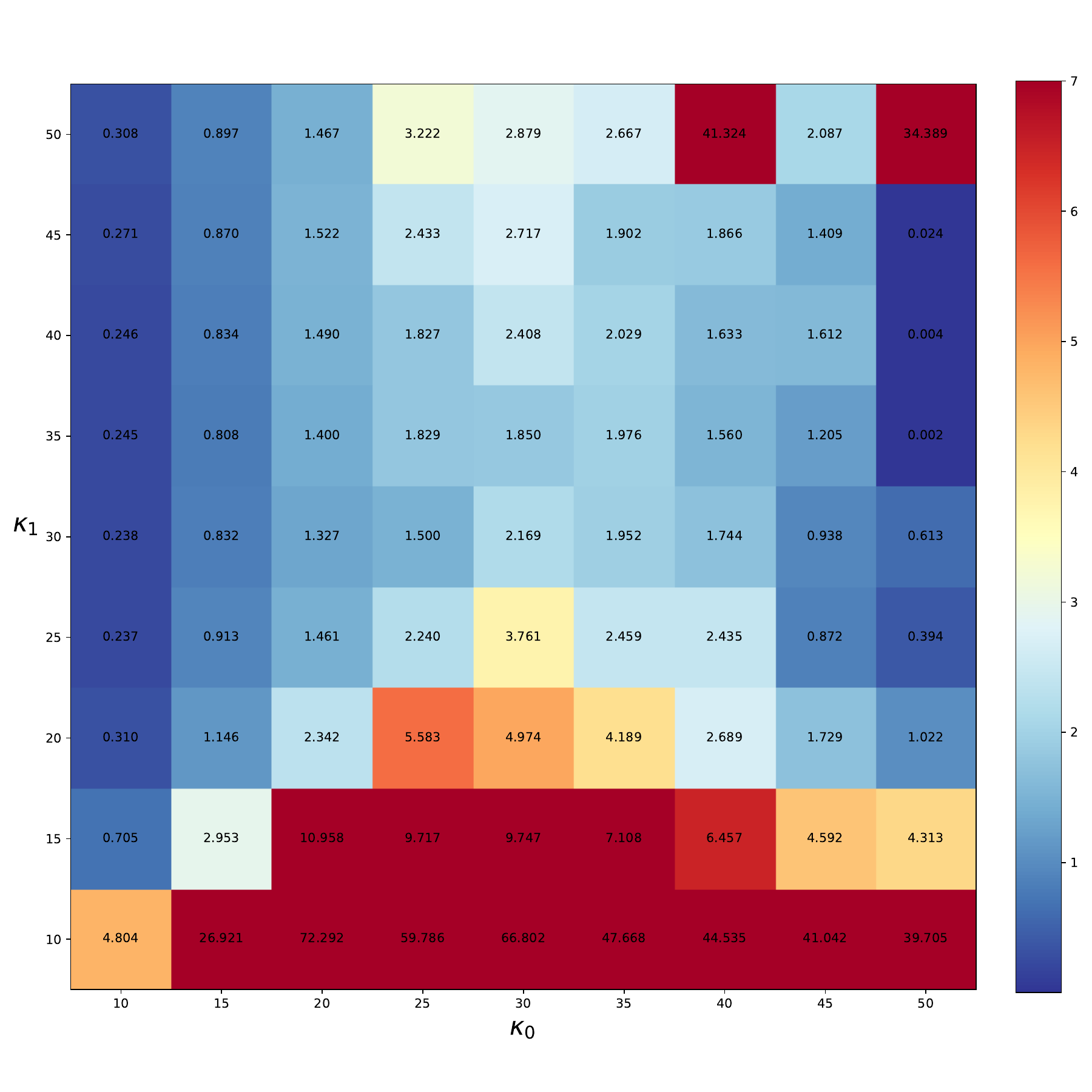}\label{fig:sub2}}
    \subfigure[$\mathrm{FRR}@(\mathrm{FAR}=10^{-3})$]{\includegraphics[width=80mm]{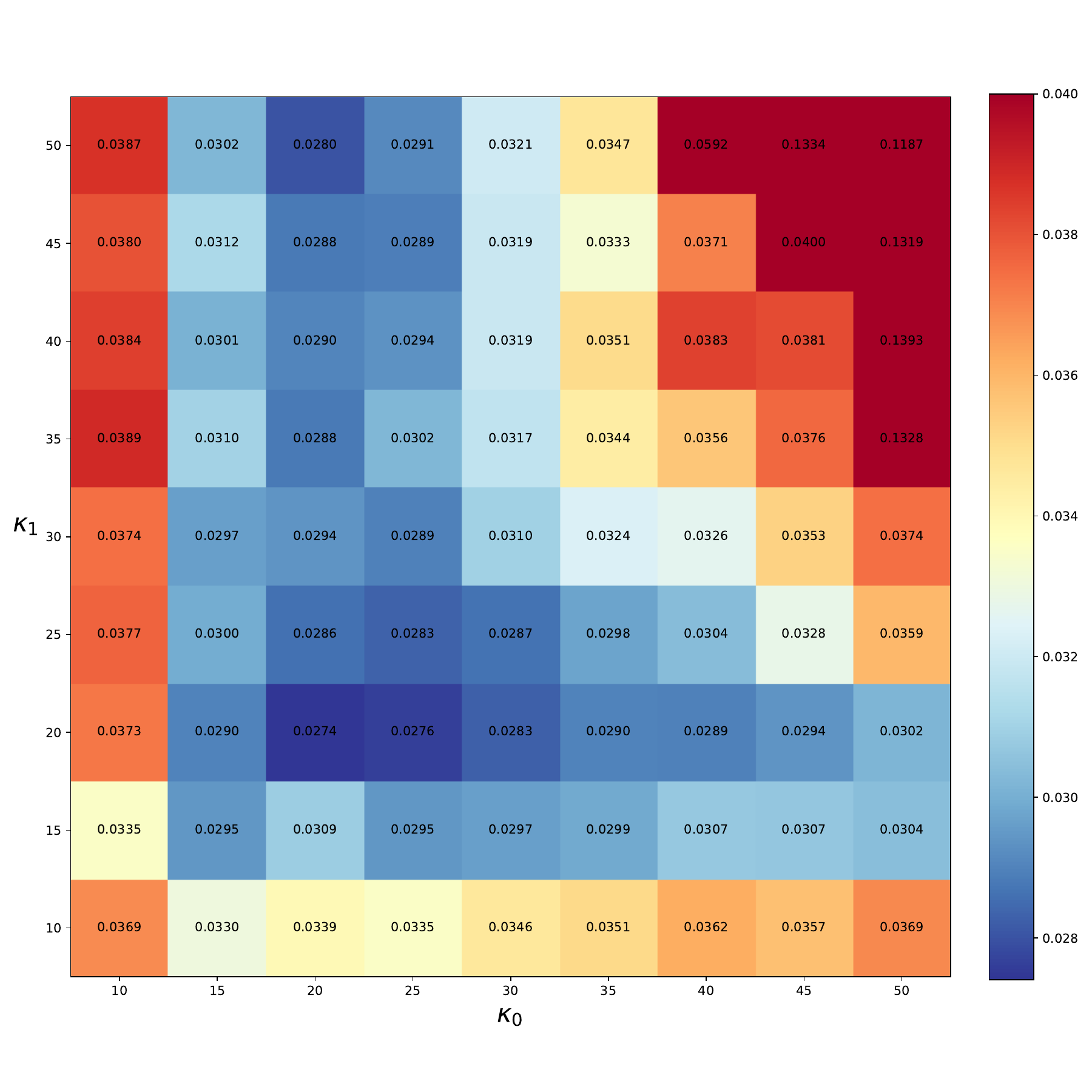}\label{fig:sub3}}
\caption{Three metrics along the grid-search. Notice that \ref{fig:sub1} and \ref{fig:sub2} are computed at the threshold $t$ satisfying $\max_{a \in \{0, 1\}} \mathrm{FAR}_a(t) = 10^{-3}$. The pre-trained model is ArcFace with a ResNet100 backbone and the Ethical Module is evaluated on IJB-C.}
\label{fig:grid_search}
\end{figure}

In order to select relevant pairs of gender-hyperparameters $(\kappa_0, \kappa_1)$, we perform a grid-search on a square of size $9 \times 9$ and keep track of the canonical performance metric $\mathrm{FRR}@(\mathrm{FAR}=10^{-3})$ together with variants of our two fairness metrics $\mathrm{BFRR}(10^{-3})$ and $\mathrm{BFAR}(10^{-3})$ introduced in Eq.~\ref{eq:fairness_metric1} and \ref{eq:fairness_metric2}. These variants are respectively $\mathrm{FRR}_1(t) / \mathrm{FRR}_0(t)$ and $\mathrm{FAR}_1(t) / \mathrm{FAR}_0(t)$ computed at the threshold $t$ satisfying $\max_{a \in \{0, 1\}} \mathrm{FAR}_a(t) = 10^{-3}$. In this way we can visualize the inversion of bias incurred by our model: in most settings, females are disadvantaged while some extreme values of the hyperparameters disadvantage males. The results displayed in Figure \ref{fig:grid_search} contain the results of Figure \ref{fig:hyperparam_trends} but they are more complete.

\section{Trends in \autoref{fig:hyperparam_trends}}\label{app:explain_trends_grid_search}

Recall that the vMF parametric interpretation of the model is that each identity is associated with a gaussian on the sphere with fixed mean and fixed concentration parameter. The images of a dataset are then seen as i.i.d. realization of the mixture of these gaussians and the loss consists in maximizing the log-likelihood. In order to control the representation power of males and females, we fix a concentration parameter $\kappa_0$ (resp. $\kappa_1$) for all males (resp. females). In \autoref{fig:hyperparam_trends}, we observe that the different metrics exhibit smooth behavior with respect to these hyperparameters. Let us give some insights on these phenomenons. In general, female are discriminated against so that the maximum is realized for $\mathrm{FAR_1}$: we will always place ourselves in this situation for the following heuristic reasoning, meaning that we will always assume that
\begin{equation} \label{eq:assump_heuristic}
\max ( \mathrm{FAR}_0(t), \mathrm{FAR}_1(t)  ) = \mathrm{FAR}_1(t). 
\end{equation}
Therefore, our heuristic will not take into account the observed empirical fact that, for some specific choices of hyperparmeters, male are discriminated against. We think one could push further the reasoning to include this case but restrict the scope of our explanations in order to focus on the underlying mechanisms of the vMF loss.

{\bf Restriction to the study of $\mathrm{FAR}_1(t)/\mathrm{FAR_0(t)}$.} We claim it is sufficient to focus on the evolution of $\mathrm{FAR}_1(t)/\mathrm{FAR_0(t)}$, from which the evolution of $\mathrm{FRR}_1(t)/\mathrm{FRR_0(t)}$ can be deduced, at least at the heuristic level developed here. Two cases may occur:
\begin{itemize}
    \item If $\mathrm{FAR}_1(t)/\mathrm{FAR_0(t)}$ increases, it means that there are more False Acceptance among females. From a geometric viewpoint, this means that females are more spread around than males. Therefore, there will be more False Reject among males who are more concentrated. Thus, when $\mathrm{FAR}_1(t)/\mathrm{FAR_0(t)}$ increases, $\mathrm{FRR}_1(t)/\mathrm{FRR_0(t)}$ decreases.
    \item If $\mathrm{FAR}_1(t)/\mathrm{FAR_0(t)}$ decreases, it means that there are less False Acceptance among females. From a geometric viewpoint, this means that females are more concentrated than males. Therefore, there will be less False Reject among males who are less concentrated. Thus, when $\mathrm{FAR}_1(t)/\mathrm{FAR_0(t)}$ decreases, $\mathrm{FRR}_1(t)/\mathrm{FRR_0(t)}$ increases.
\end{itemize}
These two observations are confirmed by the graphical representations of \autoref{fig:hyperparam_trends}.

\smallskip

\begin{figure}[h]
    \centering
    
    \subfigure[ When $\kappa_1$ is small, the $\mathrm{FAR}_0$ curve is not perturbed.]{\includegraphics[scale=0.4]{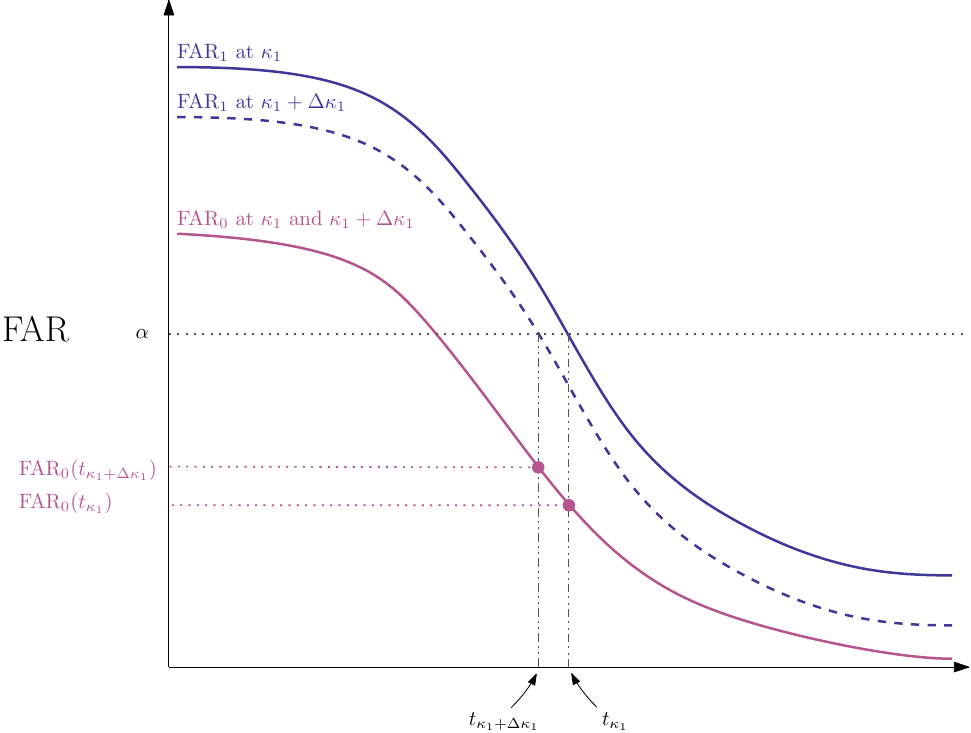}}

    \subfigure[When $\kappa_1$ is large enough, the $\mathrm{FAR}_0$ curve starts to be perturbed.]{\includegraphics[scale=0.4]{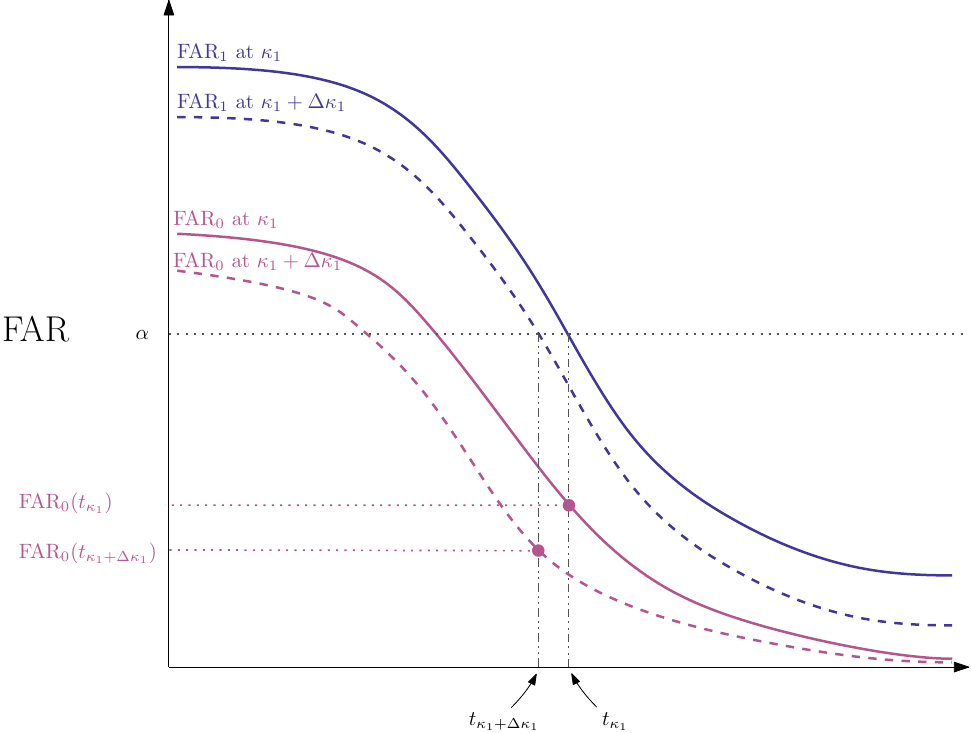}}

    
    
   \caption{Heuristic explanation of the $\mathrm{FAR}_a(t)$ evolution at fixed $\kappa_0$ and when $\kappa_1$ increases.}%
    \label{fig:hyperparam_far_0}
\end{figure}

{\bf Suppose that $\kappa_1$ is increased by a small amount $\Delta \kappa_1$ while $\kappa_0$ remains fixed.} \\
We will denote by $\mathrm{FAR}_a^{\kappa_1}$ the False Acceptance Rate curve of subgroup $a$ for the hyperparameters choice $(\kappa_0, \kappa_1)$ and by $\mathrm{FAR}_a^{\kappa_1 + \Delta \kappa_1}$ the False Acceptance Rate curve of subgroup $a$ for the hyperparameters choice $(\kappa_0, \kappa_1 + \Delta \kappa_1)$.\\
The representation with hyperparameters $(\kappa_0, \kappa_1 + \Delta \kappa_1)$ increases the concentration parameter of females. As a result, the images stemming from a same female identity should be closer from one another, leading to a better $\mathrm{FAR}$ performance. Therefore, one should have:
\begin{equation} \label{eq:far_1_kappa_1}
\forall t \in [0,1], \quad \mathrm{FAR}_1^{\kappa_1 + \Delta \kappa_1}(t) < \mathrm{FAR}_1^{\kappa_1}(t). 
\end{equation}
Let us denote by $t_{\kappa_1}$ and $t_{\kappa_1 + \Delta \kappa_1}$ the points satisfying:
\[ \mathrm{FAR}_1^{\kappa_1}(t_{\kappa_1}) = \alpha \quad \text{and} \quad \mathrm{FAR}_1^{\kappa_1 + \Delta \kappa_1}(t_{\kappa_1 + \Delta \kappa_1}) = \alpha. \]
Using \autoref{eq:assump_heuristic} and \autoref{eq:far_1_kappa_1}, this implies that $t_{\kappa_1 + \Delta \kappa_1} < t_{\kappa_1}$, as illustrated in \autoref{fig:hyperparam_far_0}.

We can now distinguish two situations depending on the magnitude of $\kappa_1$.
\begin{itemize}
    \item If $\kappa_1$ is small, its variation does not affect the representation of males at least at a first order approximation. In that case $\mathrm{FAR}^{\kappa_1}_0(t_{\kappa_1}) = \mathrm{FAR}_0^{\kappa_1 + \Delta \kappa_1}(t_{\kappa_1 + \Delta \kappa_1})$. Since $\mathrm{FAR}^{\kappa_1}_0$ is nonincreasing, we deduce that $\mathrm{FAR}_0(t_{\kappa_1 + \Delta \kappa_1}) > \mathrm{FAR}_0(t_{\kappa_1})$, which finally implies that:
\[ \frac{\mathrm{FAR}^{\kappa_1}_1(t_{\kappa_1})}{\mathrm{FAR}^{\kappa_1}_0(t_{\kappa_1})} = \frac{\alpha}{\mathrm{FAR}^{\kappa_1}_0(t_{\kappa_1})} > \frac{\alpha}{\mathrm{FAR}_0^{\kappa_1 + \Delta \kappa_1}(t_{\kappa_1 + \Delta \kappa_1})} = \frac{\mathrm{FAR}^{\kappa_1 + \Delta \kappa_1}_1(t_{\kappa_1 + \Delta \kappa_1})}{\mathrm{FAR}^{\kappa_1 + \Delta \kappa_1}_0(t_{\kappa_1 + \Delta \kappa_1})}.  \] 

    \item If $\kappa_1$ is large enough, tightening the representations of females among themselves starts to affect the males representation. Indeed, they enjoy more space and can therefore be better spread, which implies that $\mathrm{FAR}^{\kappa_1}_0(t_{\kappa_1}) > \mathrm{FAR}_0^{\kappa_1 + \Delta \kappa_1}(t_{\kappa_1 + \Delta \kappa_1})$, as illustrated in \autoref{fig:hyperparam_far_0} (b). As a result:
    \[ \frac{\mathrm{FAR}^{\kappa_1}_1(t_{\kappa_1})}{\mathrm{FAR}^{\kappa_1}_0(t_{\kappa_1})} = \frac{\alpha}{\mathrm{FAR}^{\kappa_1}_0(t_{\kappa_1})} < \frac{\alpha}{\mathrm{FAR}_0^{\kappa_1 + \Delta \kappa_1}(t_{\kappa_1 + \Delta \kappa_1})} = \frac{\mathrm{FAR}^{\kappa_1 + \Delta \kappa_1}_1(t_{\kappa_1 + \Delta \kappa_1})}{\mathrm{FAR}^{\kappa_1 + \Delta \kappa_1}_0(t_{\kappa_1 + \Delta \kappa_1})}.  \] 
\end{itemize}
The two previous points are confirmed by the top left corner graphical representation of \autoref{fig:hyperparam_trends}: for all fixed values of $\kappa_0$, the curves start by decreasing when $\kappa_1$ increases, then begin an increasing phase when $\kappa_1$ becomes sufficiently large.

{\bf Suppose that $\kappa_0$ is increased by a small amount $\Delta \kappa_0$ while $\kappa_0$ remains fixed.}\\
We will denote by $\mathrm{FAR}_a^{\kappa_0}$ the False Acceptance Rate curve of subgroup $a$ for the hyperparameters choice $(\kappa_0, \kappa_1)$ and by $\mathrm{FAR}_a^{\kappa_0 + \Delta \kappa_0}$ the False Acceptance Rate curve of subgroup $a$ for the hyperparameters choice $(\kappa_0 + \Delta \kappa_0, \kappa_1 )$.\\
The representation with hyperparameters $(\kappa_0 + \Delta \kappa_0, \kappa_1 )$ increases the concentration parameter of males. As a result, the images stemming from a same male identity should be closer from one another, leading to a better $\mathrm{FAR}$ performance. Therefore, one should have:
\begin{equation} \label{eq:far_1_kappa_0}
\forall t \in [0,1], \quad \mathrm{FAR}_0^{\kappa_0 + \Delta \kappa_0}(t) < \mathrm{FAR}_0^{\kappa_0}(t). 
\end{equation}
As before, we can distinguish two situations depending on the magnitude of $\kappa_0$.

\begin{figure}[h]
    \centering
    
    \subfigure[When $\kappa_1$ does not increase to much, the $\mathrm{FAR}_0$ curve is not perturbed.]{\includegraphics[scale=0.4]{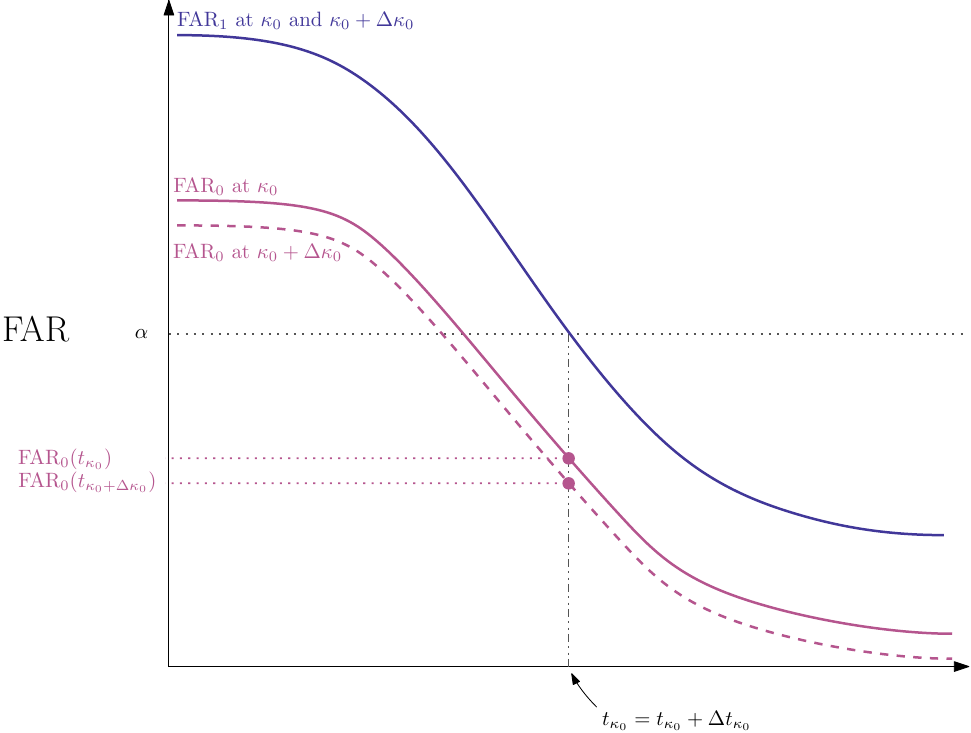}} \hspace{0.2cm}

    \subfigure[When $\kappa_1$ is large enough, the $\mathrm{FAR}_0$ curve starts to be perturbed.]{\includegraphics[scale=0.4]{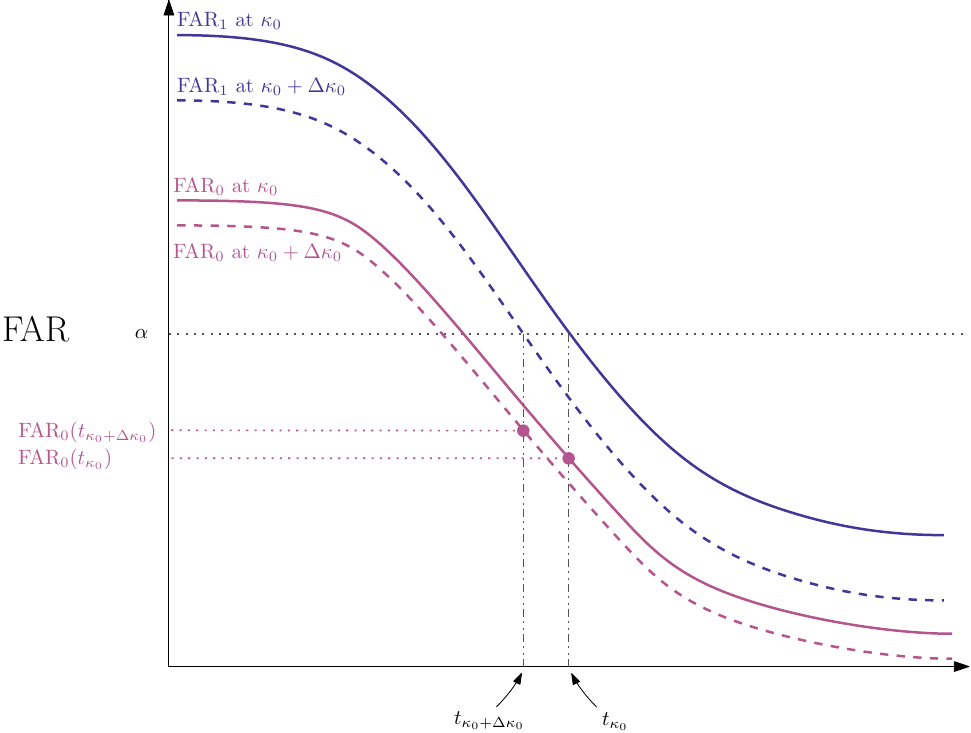}}

   \caption{Heuristic explanation of the $\mathrm{FAR}_a(t)$ evolution at fixed $\kappa_0$ and when $\kappa_1$ increases.}    \label{fig:hyperparam_farbis}
\end{figure}

\begin{itemize}
    \item If $\kappa_0$ is small, one can suppose that females are not affected by its variation, meaning that $\mathrm{FAR}_1^{\kappa_0} = \mathrm{FAR}_1^{\kappa_0 + \Delta \kappa_0}$ at a first order approximation (see (a) of \autoref{fig:hyperparam_farbis} for an illustration). In that case, $t_{\kappa_0} = t_{\kappa_0 + \Delta \kappa_0}$, and \autoref{eq:far_1_kappa_0} implies that $\mathrm{FAR}^{\kappa_0 + \Delta \kappa_0}_0(t_{\kappa_0}) < \mathrm{FAR}^{\kappa_0}_0(t_{\kappa_0}) $. As a result:
    \[ \frac{\mathrm{FAR}^{\kappa_0}_1(t_{\kappa_0})}{\mathrm{FAR}^{\kappa_0}_0(t_{\kappa_0})} = \frac{\alpha}{\mathrm{FAR}^{\kappa_0}_0(t_{\kappa_0})} < \frac{\alpha}{\mathrm{FAR}_0^{\kappa_0 + \Delta \kappa_0}(t_{\kappa_0 + \Delta \kappa_0})} = \frac{\mathrm{FAR}^{\kappa_0 + \Delta \kappa_0}_1(t_{\kappa_0 + \Delta \kappa_0})}{\mathrm{FAR}^{\kappa_0 + \Delta \kappa_0}_0(t_{\kappa_0 + \Delta \kappa_0})}.  \] 
    
    \item If $\kappa_0$ is large enough, tightening the representations of males among themselves starts to affect the females representation: they have more space to spread around (see (b) of \autoref{fig:hyperparam_farbis}). As a result, one can have $\mathrm{FAR}^{\kappa_0 + \Delta \kappa_0}_0(t_{\kappa_0 + \Delta \kappa_0}) > \mathrm{FAR}^{\kappa_0}_0(t_{\kappa_0}) $
    \[ \frac{\mathrm{FAR}^{\kappa_0}_1(t_{\kappa_0})}{\mathrm{FAR}^{\kappa_0}_0(t_{\kappa_0})} = \frac{\alpha}{\mathrm{FAR}^{\kappa_0}_0(t_{\kappa_0})} > \frac{\alpha}{\mathrm{FAR}_0^{\kappa_0 + \Delta \kappa_0}(t_{\kappa_0 + \Delta \kappa_0})} = \frac{\mathrm{FAR}^{\kappa_0 + \Delta \kappa_0}_1(t_{\kappa_0 + \Delta \kappa_0})}{\mathrm{FAR}^{\kappa_0 + \Delta \kappa_0}_0(t_{\kappa_0 + \Delta \kappa_0})}.  \] 
\end{itemize}

\section{Robustness of the selected hyperparameters}\label{app:robustness_kappas}

\begin{figure*}[ht]
\vspace{-0.2cm}
\hspace{-1cm}
    \includegraphics[scale=0.33]{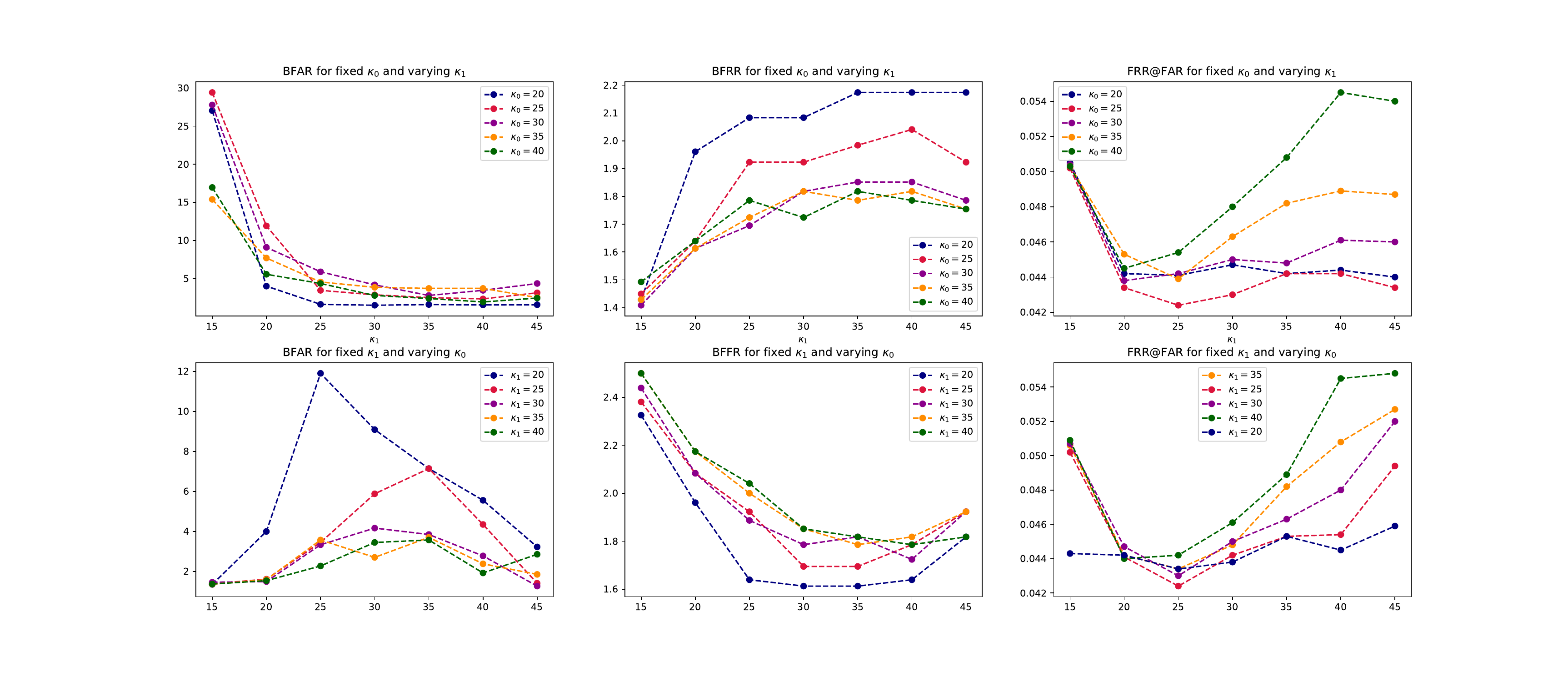}
    \vspace{-0.9cm}
    \caption{Fairness and evaluation metrics on IJB-C for the Ethical Module when one of the two hyperparameters is fixed. The FAR level defining the threshold $t$ is set to $10^{-4}$; the pre-trained model is ArcFace with a ResNet100 backbone. $\mathrm{FRR}@\mathrm{FAR}$ is expressed as a percentage (\%). The three versions of the Ethical Module presented in \autoref{tab:kappa_choice} are robust to a change of FAR level, when performing the grid-search.}
    \label{fig:hyperparam_trends2}
    \vspace{-0.1in}
\end{figure*}

The grid-search, presented in Figure \ref{fig:hyperparam_trends}, is performed using the IJB-C dataset, at a FAR level equal to $10^{-3}$.  The three versions of the Ethical Module presented in \autoref{tab:kappa_choice} are found, based on this grid-search. One relevant issue about this method is the robustness of the three selected hyperparameters, typically when performing the grid-search at a different FAR level. Figure \ref{fig:hyperparam_trends2} displays the same grid-search on IJB-C than Figure \ref{fig:hyperparam_trends}, but at a FAR level equal to $10^{-4}$. The three versions of the Ethical Module are robust to a change of FAR level on the validation set.

\section{Comparison of the spread of embeddings between genders}

The Ethical Module has hyperparameters which partly control the intra-class variance per gender. Our solution EM-FRR significantly reduces the $\mathrm{BFRR}$ metric: at any given operating point $t$, we should thus have $\mathrm{FRR}_0(t) \sim \mathrm{FRR}_1(t)$. This could be understood as follows: genuine male images are as spread around their centroid than genuine female images are around their own centroid. We would like to check whether this phenomenon occurs.

Since the training phase is an iterative process, the centroids might not represent the exact center of each identity within the hypersphere. We choose to compute the center of a given identity by the mean of the embeddings that form this identity, renormalized to lie on the hypersphere.

To measure the variability of the embeddings $z_1, \ldots, z_n$ of a given identity, we first compute the empirical mean $\overline{z} := (1/n) \sum_{1 \leq i \leq n} z_i$, and renormalize it on the hypersphere: $\mathfrak{z} = \overline{z} / || \overline{z} ||_2$. Then, we compute the inertia of $z_1, \ldots, z_n$ with respect to $\mathfrak{z}$:

\[ I := \frac{1}{n} \sum\limits_{i=1}^n || z_i - \mathfrak{z} ||_2^2. \]

We use the pre-trained model ArcFace ResNet100 for this experiment; trained on MS1MV3 as our EM-FRR. In order to have good estimates of this spread measure, we only considered identities having at least $100$ images within the training set (MS1MV3). We end up with $3569$ male identities and $5045$ female identities. For each of those identities, we compute the spread measure and get an histogram of those values per gender (one histogram for male identities, another for female identities).

\begin{figure}[h]
    \centering
    
    \subfigure[ArcFace]{\includegraphics[scale=0.4]{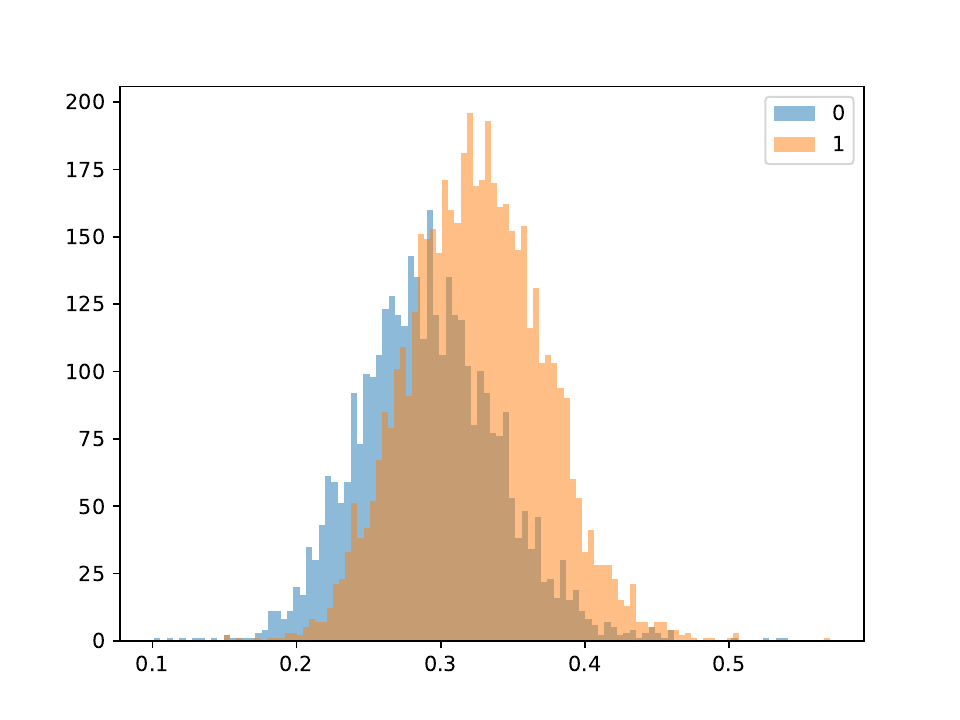}}
    \caption{Histograms of identities inertias. In orange: for females. In blue: for males. For the pretrained model, the two histograms are not aligned.}
    \hspace{0.2cm}

    \subfigure[EM-FRR]{\includegraphics[scale=0.4]{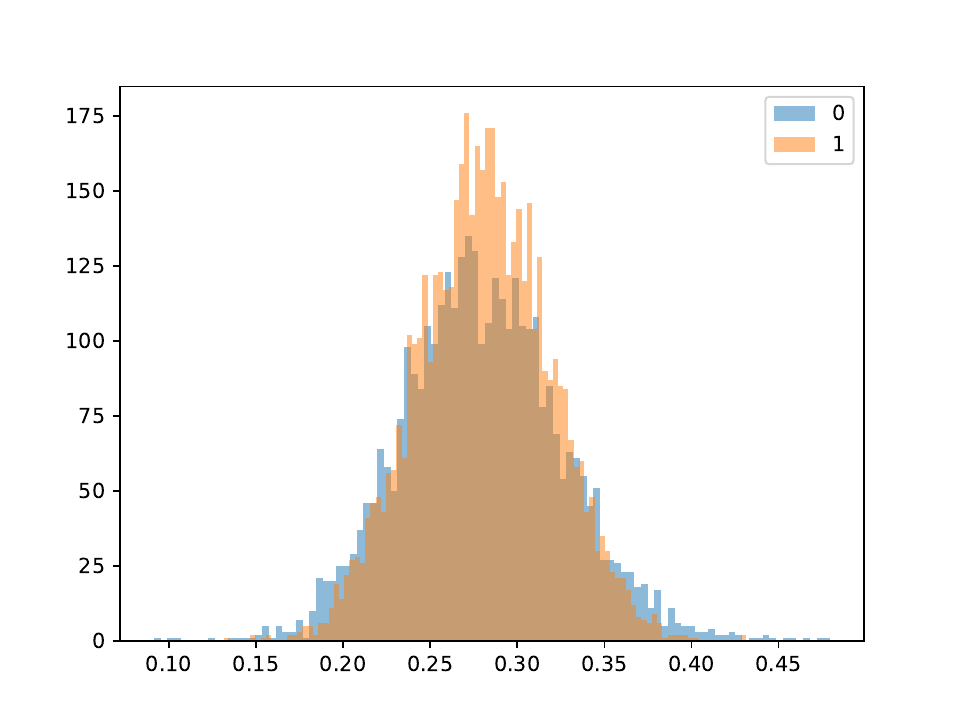}}
    \caption{Histograms of identities inertias. In orange: for females. In blue: for males. For our EM-FRR model, the two histograms are aligned.}    \label{}
\end{figure}

\section{Influence of $\kappa_1$ on $\mathrm{FAR}_a(t)$ and $\mathrm{FRR}_a(t)$}

\begin{figure}[h]
    \centering
    \subfigure[Influence of $\kappa_1$ on $\mathrm{FAR}_a(t)$. Females are depicted with solid lines while males are depicted with dashed lines. ]{\includegraphics[scale=0.6]{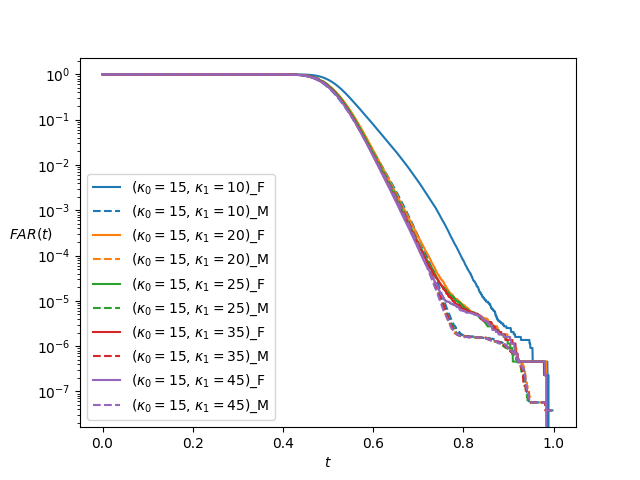}\label{fig:kappa_f_influence_FAR_a(t)_all}} \hspace{2cm}
    \subfigure[Zoom on Figure \ref{fig:kappa_f_influence_FAR_a(t)_all} for females.]{\includegraphics[scale=0.4]{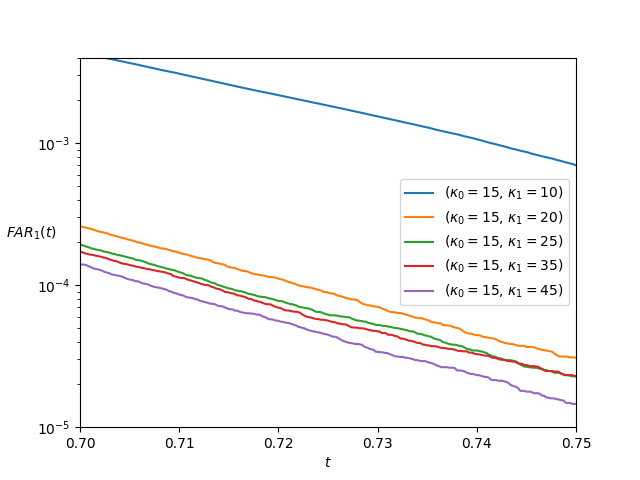}}
    \subfigure[Same as Figure \ref{fig:kappa_f_influence_FAR_a(t)_all} but only males are displayed. The female concentration parameter does not affect $\mathrm{FAR}_0(t)$.]{\includegraphics[scale=0.4]{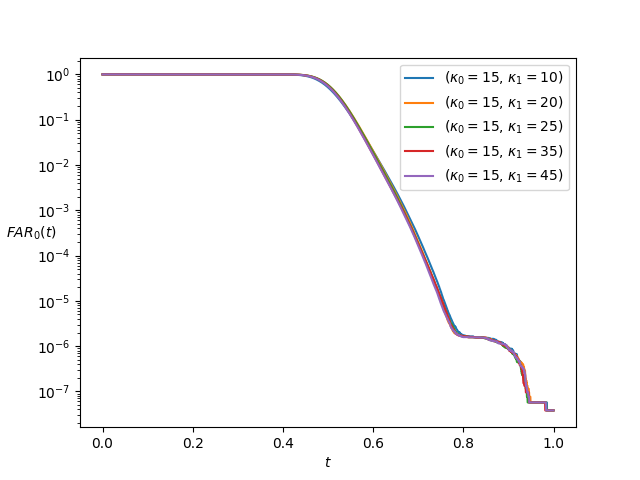}}
   \caption{Influence of $\kappa_1$ on $\mathrm{FAR}_a(t)$.}    \label{fig:kappa_f_influence_FAR_a(t)}
\end{figure}

\begin{figure}[h]
    \centering
    \subfigure[Influence of $\kappa_1$ on $\mathrm{FRR}_a(t)$. Females are depicted with solid lines while males are depicted with dashed lines. ]{\includegraphics[scale=0.6]{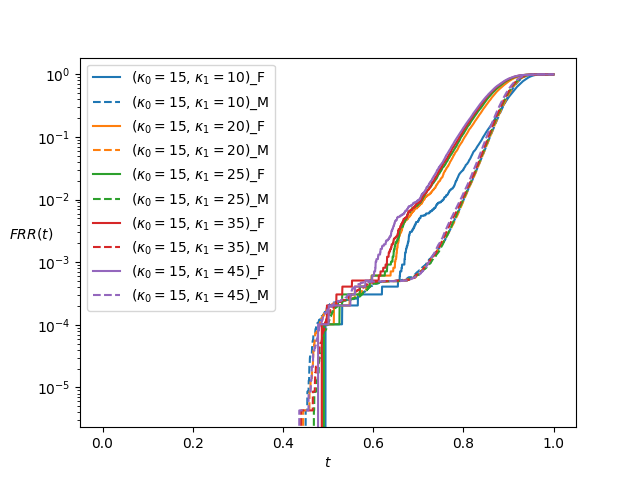}\label{fig:kappa_f_influence_FRR_a(t)_all}} \hspace{2cm}
    \subfigure[Zoom on Figure \ref{fig:kappa_f_influence_FRR_a(t)_all} for females.]{\includegraphics[scale=0.4]{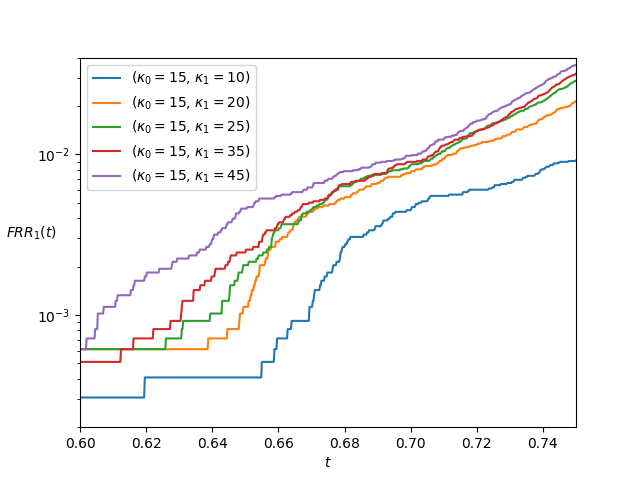}}
    \subfigure[Same as Figure \ref{fig:kappa_f_influence_FRR_a(t)_all} but only males are displayed. The female concentration parameter barely affects $\mathrm{FRR}_0(t)$.]{\includegraphics[scale=0.4]{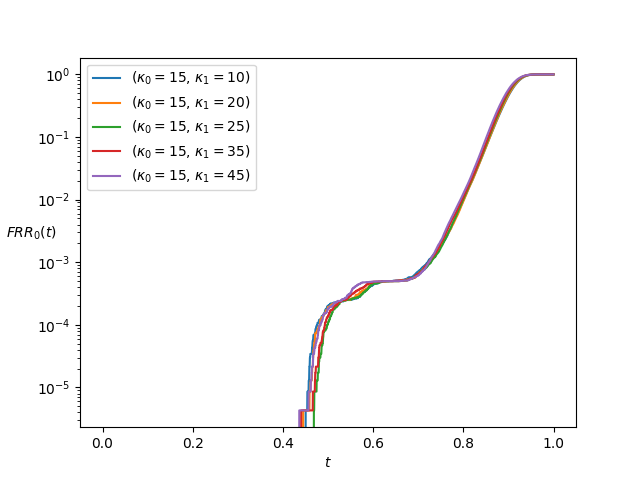}}
   \caption{Influence of $\kappa_1$ on $\mathrm{FRR}_a(t)$.}    \label{fig:kappa_f_influence_FRR_a(t)}
\end{figure}

After training our Ethical Module with ArcFace ResNet100 as the pre-trained model, we evaluate it on the LFW dataset and compute the quantities $\mathrm{FAR}_a(t)$ (Figure \ref{fig:kappa_f_influence_FAR_a(t)}) and $\mathrm{FRR}_a(t)$ (Figure \ref{fig:kappa_f_influence_FRR_a(t)}) with varying $\kappa_1$. This shows that our vMF mixture statistical model has a clear impact on the representation of deep embeddings.

\clearpage

\section{Additional numerical results}\label{app:additional_results}

In this section, we provide more numerical experiments, varying the evaluation dataset (LFW, IJB-C, IJB-B) and different kinds of pre-trained models (ArcFace with several ResNet architectures, other pre-trained models with MobileFaceNet backbone).


\subsection{Fairness evaluation on IJB-C and LFW}

In \autoref{tab:table4}, \autoref{tab:table5}, \autoref{tab:table6}, \autoref{tab:table7}, we provide additional fairness evaluations on the IJB-C and LFW datasets, for ArcFace (with different ResNet architectures) and other pre-trained models with MobileFaceNet backbone (CosFace, CurricularFace, AdaCos~\cite{adacos}).

\begin{table*}
\center
\caption{Evaluation on LFW for ArcFace with ResNet100 backbone and different pre-trained models (AdaCos, CosFace, CurricularFace) with MobileFaceNet backbone. By "original" we mean no Ethical Module is added to the pre-trained model. The tuples correspond to the choices of $\kappa_0$ (first argument) and $\kappa_1$ (second argument). $\mathrm{FRR}@\mathrm{FAR}$ is expressed as a percentage (\%).}
\begin{tabular}{ c | c | ccc | ccc}
 \multicolumn{2}{c}{ $\mathrm{FAR}$ level: }         & \multicolumn{3}{c}{ $10^{-4}$} & \multicolumn{3}{c}{ $10^{-3}$} \\ 
 \hline \hline
  \multicolumn{2}{c}{ model }    & $\mathrm{FRR}@\mathrm{FAR}$ (\%)  & $\mathrm{BFRR}$  & $\mathrm{BFAR}$  & $\mathrm{FRR}@\mathrm{FAR}$ (\%)  & $\mathrm{BFRR}$        & $\mathrm{BFAR}$        \\ \hline
                              & original  & $\mathbf{0.063}$  & $\underline{10.76}$ & $3.98$  & $\mathbf{0.052}$   & $\underline{2.23}$  & $1.81$ \\
  \multirow{2}{*}{ArcFace}    & (15,20) & $0.119$  & $12.73$ & $\mathbf{1.72}$  & $0.067$ & $8.43$ &  $\mathbf{1.04}$ \\
                              & (25,20) & $\underline{0.076}$ & $\mathbf{5.35}$ & $29.33$ & $\underline{0.052}$ & $\mathbf{1.94}$  & $3.96$ \\
                              & (45,30) & $0.129$  & $13.47$ & $\underline{2.99}$ & $0.067$ & $6.02$  & $\underline{1.24}$  \\ \hline   \hline   
                              & original  & $\mathbf{2.97}$ & $\underline{3.64}$ & $3.84$ & $\underline{0.98}$  & $\underline{5.29}$ & $2.23$ \\
  \multirow{2}{*}{AdaCos}     & (15,20) & $4.56$ & $4.42$ & $\mathbf{1.41}$  & $1.33$ & $6.34$ & $\mathbf{1.01}$  \\
                              & (25,20) & $\underline{3.12}$ & $\mathbf{2.71}$ & $8.37$ & $\mathbf{0.91}$ & $\mathbf{4.23}$  & $3.71$ \\
                              & (45,30) & $4.05$ & $4.51$  & $\underline{1.57}$ & $1.26$ & $7.28$ & $\underline{1.08}$  \\ \hline 
                              & original  & $\mathbf{1.73}$ & $5.89$ & $\underline{2.51}$ & $\mathbf{0.58}$  & $8.18$  & $\underline{1.74}$  \\
\multirow{2}{*}{CosFace}  & (15,20) & $3.69$ & $5.76$ & $\mathbf{1.13}$  & $1.05$ & $8.41$ &  $\mathbf{1.02}$ \\
                              & (25,20) & $\underline{2.41}$ & $\mathbf{3.03}$ & $9.66$ & $\underline{0.67}$ & $\mathbf{5.09}$  & $4.75$ \\
                              & (45,30) & $2.60$ & $\underline{4.30}$ & $3.69$ & $0.82$ & $\underline{6.81}$ & $1.87$ \\ \hline 
                              & original  & $\mathbf{2.52}$ & $3.67$ & $2.92$ & $\mathbf{0.81}$  & $\underline{4.88}$ & $1.91$  \\
  \multirow{2}{*}{Curricular} & (15,20) & $3.86$ & $5.26$ & $\mathbf{1.16}$  & $1.17$ & $6.35$  & $\mathbf{1.10}$  \\
                              & (25,20) & $\underline{2.82}$ & $\mathbf{2.58}$ & $9.10$ & $\underline{0.82}$ & $\mathbf{3.89}$  & $4.28$ \\
                              & (45,30) & $3.61$ & $\underline{3.40}$ & $\underline{2.30}$ & $1.02$ & $5.63$ & $\underline{1.27}$  \\ \hline 
\end{tabular}
\label{tab:table4}
\end{table*}

\begin{table*}
\center
\caption{IJBC 1:1 protocol for ArcFace with ResNet100 backbone and different pre-trained models with MobileFaceNet backbone. By "original" we mean no Ethical Module is added to the pre-trained model. The tuples correspond to the choices of $\kappa_0$ (first argument) and $\kappa_1$ (second argument).}
\begin{tabular}{ c | c | ccc | ccc}
 \multicolumn{2}{c}{ FAR level: }         & \multicolumn{3}{c}{ $10^{-4}$} & \multicolumn{3}{c}{ $10^{-3}$} \\ 
 \hline \hline
  \multicolumn{2}{c}{ model }    & $\mathrm{FRR}@\mathrm{FAR}$ (\%)  & $\mathrm{BFRR}$  & $\mathrm{BFAR}$  & $\mathrm{FRR}@\mathrm{FAR}$ (\%)  & $\mathrm{BFRR}$        & $\mathrm{BFAR}$        \\ \hline
                              & original  & $\mathbf{3.94}$  & $1.97$ & $4.06$  & $\mathbf{2.68}$  & $1.79$  & $2.04$  \\
  \multirow{2}{*}{ArcFace}    & (15,20) & $4.90$  & $2.33$ & $\mathbf{1.17}$  & $2.90$  & $1.98$  & $1.15$  \\
                              & (25,20) & $4.34$  & $\mathbf{1.62}$ & $11.86$ & $2.76$  & $\mathbf{1.60}$  & $5.58$  \\
                              & (45,30) & $5.20$  & $1.92$ & $1.25$  & $3.53$  & $1.91$  & $\mathbf{1.07}$  \\ \hline   \hline   
                              & original  & $18.85$ & $1.18$ & $5.44$  & $\mathbf{9.74}$  & $1.24$  & $3.84$  \\
  \multirow{2}{*}{AdaCos}     & (15,20) & $20.75$ & $1.30$ & $\mathbf{2.06}$  & $10.31$ & $1.42$  & $2.20$  \\
                              & (25,20) & $20.28$ & $\mathbf{1.02}$ & $13.00$ & $10.09$ & $\mathbf{1.06}$  & $7.80$  \\
                              & (45,30) & $\mathbf{17.48}$ & $1.28$ & $2.86$  & $9.85$  & $1.33$  & $\mathbf{2.06}$  \\ \hline 
                              & original  & $\mathbf{15.67}$ & $1.24$ & $3.08$  & $\mathbf{8.55}$  & $1.35$  & $2.54$  \\
\multirow{2}{*}{CosFace}  & (15,20) & $19.52$ & $1.35$ & $\mathbf{2.75}$  & $10.24$ & $1.41$  & $\mathbf{2.33}$  \\
                              & (25,20) & $20.57$ & $\mathbf{1.03}$ & $86.69$ & $10.24$ & $\mathbf{1.04}$  & $13.61$ \\
                              & (45,30) & $17.27$ & $1.12$ & $8.67$  & $9.69$  & $1.11$  & $4.29$  \\ \hline 
                              & original  & $\mathbf{17.69}$ & $1.19$ & $8.18$  & $\mathbf{9.26}$  & $1.30$  & $4.21$  \\
  \multirow{2}{*}{Curricular} & (15,20) & $19.97$ & $1.33$ & $\mathbf{3.23}$  & $10.37$ & $1.42$  & $\mathbf{2.23}$  \\
                              & (25,20) & $20.35$ & $\mathbf{1.04}$ & $20.88$ & $10.02$ & $\mathbf{1.03}$  & $9.54$  \\
                              & (45,30) & $18.07$ & $1.18$ & $5.29$  & $9.99$  & $1.22$  &  $3.33$ \\ \hline 
\end{tabular}
\label{tab:table5}
\end{table*}

\begin{table*}
\center
\caption{Evaluation on LFW for ArcFace on different ResNet architectures. By "original" we mean no Ethical Module is added to the pre-trained model. The tuples correspond to the choices of $\kappa_0$ (first argument) and $\kappa_1$ (second argument).}
\begin{tabular}{ c | c | ccc | ccc}
 \multicolumn{2}{c}{ FAR level: }         & \multicolumn{3}{c}{ $10^{-4}$} & \multicolumn{3}{c}{ $10^{-3}$} \\ 
 \hline \hline
  \multicolumn{2}{c}{model}    & $\mathrm{FRR}@\mathrm{FAR}$ (\%)  & $\mathrm{BFRR}$  & $\mathrm{BFAR}$  & $\mathrm{FRR}@\mathrm{FAR}$ (\%)  & $\mathrm{BFRR}$        & $\mathrm{BFAR}$        \\ \hline
                              & original  & $\mathbf{0.063}$ & $10.76$ & $3.98$  & $\mathbf{0.052}$  & $2.23$ & $1.81$  \\
  \multirow{2}{*}{R100}        & (15,20) & $0.119$ & $12.73$ & $\mathbf{1.72}$ & $0.067$  & $8.43$ &  $\mathbf{1.04}$ \\
                              & (25,20) & $0.076$ & $\mathbf{5.35}$ & $29.33$ & $0.052$ & $\mathbf{1.94}$ & $3.96$  \\
                              & (45,30) & $0.129$ & $13.47$ & $2.99$ & $0.067$  & $6.02$ & $1.24$  \\ \hline     
                              & original  & $\mathbf{0.078}$ & $10.27$ & $4.72$ & $0.059$  & $4.17$ & $1.81$ \\
  \multirow{2}{*}{R50}        & (15,20) & $0.151$ & $11.22$ & $\mathbf{2.11}$ & $0.072$  & $9.16$ & $\mathbf{1.19}$  \\
                              & (25,20) & $0.100$ & $\mathbf{5.89}$ & $33.65$ & $\mathbf{0.058}$ & $\mathbf{4.11}$ & $5.24$  \\
                              & (45,30) & $0.164$ & $9.18$ & $2.44$ & $0.081$  & $5.15$ & $1.20$  \\ \hline 
                              & original  & $\mathbf{0.104}$ & $11.81$ & $7.62$ & $\mathbf{0.063}$  & $8.64$ & $2.17$ \\
\multirow{2}{*}{R34}          & (15,20) & $0.204$ & $14.27$ & $\mathbf{3.31}$ & $0.087$  & $17.56$ & $1.59$  \\
                              & (25,20) & $0.163$ & $\mathbf{5.63}$ & $43.55$ & $0.069$  & $\mathbf{8.09}$ & $6.43$ \\
                              & (45,30) & $0.226$ & $8.85$ & $4.42$ & $0.095$  & $8.80$ & $\mathbf{1.02}$  \\ \hline 
                              & original  & $\mathbf{0.214}$ & $11.16$ & $2.80$ & $\mathbf{0.116}$  & $7.53$ & $1.93$ \\
  \multirow{2}{*}{R18}        & (15,20) & $0.465$ & $11.15$ & $\mathbf{1.59}$ & $0.197$  & $10.60$ & $\mathbf{1.34}$  \\
                              & (25,20) & $0.310$ & $\mathbf{4.44}$ & $24.59$ & $0.125$  & $\mathbf{6.53}$ & $7.57$ \\
                              & (45,30) & $0.349$ & $6.69$ & $4.21$ & $0.162$  & $6.92$ & $1.76$  \\ \hline 
\end{tabular}
\label{tab:table6}
\end{table*}

\begin{table}
\center
\caption{IJBC 1:1 protocol for ArcFace on different ResNet architectures. By "original" we mean no Ethical Module is added to the pre-trained model. The tuples correspond to the choices of $\kappa_0$ (first argument) and $\kappa_1$ (second argument). Notice that PASS-g performs as well as our method on BFRR, but at the price of a poor performance / BFAR metrics compared to our method.}
\begin{tabular}{ c | c | ccc | ccc}
 \multicolumn{2}{c}{ FAR level: }         & \multicolumn{3}{c}{ $10^{-4}$} & \multicolumn{3}{c}{ $10^{-3}$} \\ 
 \hline \hline
  \multicolumn{2}{c}{model}    & $\mathrm{FRR}@\mathrm{FAR}$ (\%)  & $\mathrm{BFRR}$  & $\mathrm{BFAR}$  & $\mathrm{FRR}@\mathrm{FAR}$ (\%)  & $\mathrm{BFRR}$        & $\mathrm{BFAR}$        \\ \hline
                              & original  & $\mathbf{3.94}$ & $1.97$ & $4.06$ &  $\mathbf{2.68}$ & $1.79$ & $2.04$  \\
  \multirow{2}{*}{R100}        & (15,20) & $4.90$ & $2.33$ & $\mathbf{1.17}$ &  $2.90$ & $1.98$ & $\underline{1.15}$  \\
                              & (25,20) & $\underline{4.34}$ & $\mathbf{1.62}$ & $11.86$&  $\underline{2.76}$ & $\mathbf{1.60}$ & $5.58$  \\
                              & (45,30) & $5.20$ & $1.92$ & $\underline{1.25}$ &  $3.53$ & $1.91$ & $\mathbf{1.07}$  \\ 
                              & PASS-g & $9.00$ & $\underline{1.70}$ & $4.49$ &  $6.27$ & $\underline{1.79}$ & $2.97$  \\ \hline     
                              & original  & $\mathbf{4.29}$ & $1.81$ & $3.41$ &  $\mathbf{3.00}$ & $1.88$ & $1.95$  \\
  \multirow{2}{*}{R50}        & (15,20) & $5.56$ & $2.18$ & $1.28$ &  $3.40$ & $2.18$ & $\mathbf{1.00}$  \\
                              & (25,20) & $\underline{4.91}$ & $\underline{1.49}$ & $10.87$&  $\underline{3.19}$ & $\mathbf{1.50}$ & $6.49$  \\
                              & (45,30) & $5.41$ & $1.73$ & $\mathbf{1.24}$ &  $3.71$ & $1.77$ & $\underline{1.09}$  \\ 
                              & PASS-g & $10.34$ & $\mathbf{1.45}$ & $6.93$ &  $7.06$ & $\underline{1.51}$ & $3.63$  \\ \hline 
                              & original  & $\mathbf{4.95}$ & $1.72$ & $2.83$ &  $\mathbf{3.47}$ & $1.77$ & $1.88$  \\
\multirow{2}{*}{R34}          & (15,20) & $6.38$ & $2.05$ & $\mathbf{1.17}$ &  $3.85$ & $2.04$ & $\mathbf{1.06}$  \\
                              & (25,20) & $\underline{5.67}$ & $\underline{1.45}$ & $13.69$&  $\underline{3.60}$ & $\mathbf{1.50}$ & $5.86$  \\
                              & (45,30) & $6.13$ & $1.62$ & $\underline{1.70}$ &  $4.24$ & $1.69$ & $\mathbf{1.06}$  \\
                              & PASS-g & $12.03$ & $\mathbf{1.43}$ & $4.00$ &  $8.36$ & $\mathbf{1.50}$ & $2.79$  \\ \hline 
                              & original  & $\mathbf{6.64}$ & $1.68$ & $3.81$ &  $\mathbf{4.41}$ & $1.58$ & $2.37$  \\
  \multirow{2}{*}{R18}        & (15,20) & $8.64$ & $1.83$ & $\mathbf{1.39}$ &  $4.96$ & $1.88$ & $\mathbf{1.43}$  \\
                              & (25,20) & $8.27$ & $\mathbf{1.19}$ & $16.25$&  $4.76$ & $\mathbf{1.26}$ & $10.94$ \\
                              & (45,30) & $7.46$ & $1.50$ & $3.16$ &  $4.97$ & $1.56$ & $1.85$  \\ \hline 
\end{tabular}
\label{tab:table7}
\end{table}

\subsection{Verification evaluation on IJB-B}

We finally investigate the $\mathrm{FRR}@\mathrm{FAR}$ metric (Table~\ref{tab:table8}, Table~\ref{tab:table9}) of the three selected points ($\kappa_0, \kappa_1$) on IJB-B \cite{whitelam2017iarpa}. In the verification setting, this dataset contains $10$k genuine pairs and $8$M impostor pairs. Notice that we do not lose too much in performance with respect to the original model.


\begin{table}
\centering
\caption{IJB-B 1:1 protocol for ArcFace on different ResNet architectures. By "original" we mean no Ethical Module is added to the pre-trained model. The tuples correspond to the choices of $\kappa_0$ (first argument) and $\kappa_1$ (second argument).}
\begin{tabular}{ c | c | c | c}

 & & \multicolumn{2}{c}{$\mathrm{FRR}@\mathrm{FAR}$ (\%)} \\ \hline \hline
 \multicolumn{2}{c}{ FAR level: }         & $10^{-4}$ & $10^{-3}$ \\  
 \hline \hline
                              & original  & $5.38$ & $3.78$  \\
  \multirow{2}{*}{R100}       & (15,20) & $6.79$ & $4.11$  \\
                              & (25,20) & $6.00$ & $3.84$  \\
                              & (45,30) & $7.03$ & $4.81$  \\ \hline     
                              & original  & $5.95$ & $4.20$  \\
  \multirow{2}{*}{R50}        & (15,20) & $7.58$ & $4.71$  \\
                              & (25,20) & $6.71$ & $4.26$  \\
                              & (45,30) & $7.34$ & $5.10$  \\ \hline 
                              & original  & $6.72$ & $4.63$  \\
\multirow{2}{*}{R34}          & (15,20) & $8.54$ & $5.18$  \\
                              & (25,20) & $7.62$ & $4.60$  \\
                              & (45,30) & $8.11$ & $5.57$  \\ \hline 
                              & original  & $8.59$ & $5.76$  \\
  \multirow{2}{*}{R18}        & (15,20) & $11.12$& $6.53$  \\
                              & (25,20) & $10.94$& $6.01$ \\
                              & (45,30) & $9.72$ & $6.35$  \\ \hline 
\end{tabular}
\label{tab:table8}
\end{table}

\begin{table}
\centering
\caption{IJB-B 1:1 protocol for ArcFace with ResNet100 backbone and different pre-trained models with MobileFaceNet backbone. By "original" we mean no Ethical Module is added to the pre-trained model. The tuples correspond to the choices of $\kappa_0$ (first argument) and $\kappa_1$ (second argument).}
\begin{tabular}{ c | c | c | c}

 & & \multicolumn{2}{c}{$\mathrm{FRR}@\mathrm{FAR}$ (\%)} \\ \hline \hline
 \multicolumn{2}{c}{ FAR level: }         & $10^{-4}$ & $10^{-3}$ \\  
 \hline \hline
                              & original  & $5.38$ & $3.78$  \\
  \multirow{2}{*}{ArcFace}    & (15,20) & $6.79$ & $4.11$  \\
                              & (25,20) & $6.00$ & $3.84$  \\
                              & (45,30) & $7.03$ & $4.81$  \\ \hline   \hline   
                              & original  & $22.98$& $12.27$  \\
  \multirow{2}{*}{AdaCos}     & (15,20) & $24.06$& $12.78$  \\
                              & (25,20) & $24.41$& $12.78$  \\
                              & (45,30) & $21.25$& $12.44$  \\ \hline 
                              & original  & $18.85$& $10.65$  \\
\multirow{2}{*}{CosFace}      & (15,20) & $23.38$& $12.51$  \\
                              & (25,20) & $26.10$& $13.01$  \\
                              & (45,30) & $21.22$& $12.27$  \\ \hline 
                              & original  & $12.20$& $11.42$  \\
  \multirow{2}{*}{Curricular} & (15,20) & $24.50$& $12.56$  \\
                              & (25,20) & $24.91$& $12.35$  \\
                              & (45,30) & $21.88$& $11.97$  \\ \hline 
\end{tabular}
\label{tab:table9}
\end{table}

\end{document}